\documentclass{article} 
\usepackage[final]{colm2026_conference}
\usepackage{pifont}
\usepackage{threeparttable}
\usepackage{multirow}
\usepackage{ifthen}
\usepackage{caption}
\usepackage{colortbl}
\usepackage{wrapfig}
\usepackage{tcolorbox}
\usepackage[table]{xcolor}
\usepackage{xcolor}
\definecolor{myyellow}{HTML}{FFCB64}
\definecolor{myteal}{HTML}{62D4C5}
\definecolor{myblue}{HTML}{3266E3}
\definecolor{mypink}{HTML}{FE839C}

\definecolor{textblue}{HTML}{3A86FF}
\definecolor{mygreen}{HTML}{219653}

\definecolor{mygreen}{HTML}{166534}

\makeatletter
\renewcommand{\fnum@figure}{\figurename~\textcolor{violet}{\thefigure}}
\renewcommand{\fnum@table}{\tablename~\textcolor{violet}{\thetable}}

\newcommand{\squishlisttwo}[1][$\bullet$]{%
 \begin{list}{#1}{%
    \usecounter{squishlisttwoitem}%
    \setlength{\itemsep}{1pt}
    \setlength{\parsep}{0pt}
    \setlength{\topsep}{0pt}
    \setlength{\parskip}{0pt}%
    \setlength{\partopsep}{0pt}
    \setlength{\leftmargin}{1em}
    \setlength{\labelwidth}{1.5em}
    \setlength{\labelsep}{0.5em}
 }%
}
\newcommand{\squishend}{\end{list}}

\usepackage{anyfontsize}
\usepackage{microtype}
\usepackage{lineno}
\usepackage{enumitem}

\usepackage{amsmath}
\usepackage{amssymb}
\usepackage{mathtools}

\usepackage{algorithm}
\usepackage{algpseudocode}

\usepackage{booktabs}
\usepackage{array}
\usepackage{tabularx}
\usepackage{makecell}

\usepackage{graphicx}
\usepackage{subcaption}
\usepackage{caption}

\usepackage{hyperref}
\usepackage{url}

\definecolor{darkblue}{rgb}{0, 0, 0.5}
\hypersetup{colorlinks=true, citecolor=darkblue, linkcolor=darkblue, urlcolor=darkblue}

\newcommand{\stage}[1]{\textsc{#1}}

\graphicspath{{figures/}}

\newcommand{\method}{\textsc{CORAL}}

\usepackage{tikz}
\usetikzlibrary{arrows.meta}

\usepackage{tcolorbox}
\tcbuselibrary{breakable,skins}

\newtcolorbox{promptbox}[1][]{%
  enhanced,
  colback=blue!3,
  colframe=blue!40!gray!60,
  fonttitle=\bfseries\small\sffamily,
  coltitle=blue!40!black,
  attach boxed title to top left={yshift=-2mm, xshift=4mm},
  boxed title style={colback=blue!10, colframe=blue!40!gray!60, boxrule=0.4pt, arc=1.5pt},
  boxrule=0.5pt,
  arc=2pt,
  left=6pt, right=6pt, top=6pt, bottom=4pt,
  fontupper=\small,
  breakable,
  title={#1},
}
\newtcolorbox{heartbeatbox}[1][]{%
  enhanced,
  colback=blue!3,
  colframe=blue!40!gray!60,
  fonttitle=\bfseries\small\sffamily,
  coltitle=blue!40!black,
  attach boxed title to top left={yshift=-2mm, xshift=4mm},
  boxed title style={colback=blue!10, colframe=blue!40!gray!60, boxrule=0.4pt, arc=1.5pt},
  boxrule=0.5pt,
  arc=2pt,
  left=6pt, right=6pt, top=6pt, bottom=4pt,
  fontupper=\small,
  breakable,
  title={#1},
}

\newcommand{\hl}[1]{\colorbox{cyan!10}{\makebox[2.8em][r]{#1}}}

\title{CORAL: Towards Autonomous Multi-Agent Evolution\\for Open-Ended Discovery}

\author{Ao Qu$^{1,8*}$ \quad Han Zheng$^{1*}$ \quad Zijian Zhou$^{2,3*}$ \quad
\textbf{Yihao Yan} \quad \textbf{Yihong Tang$^{4}$} \\[4pt] \textbf{Shao Yong Ong$^{2}$} \quad \textbf{Fenglu Hong$^{5,6}$} \quad \textbf{Kaichen Zhou$^{1}$} \quad
\textbf{Chonghe Jiang$^{1,8}$} \\[4pt]
\textbf{Minwei Kong$^{8}$} \quad \textbf{Jiacheng Zhu$^{7\ddagger}$} \quad \textbf{Xuan Jiang$^{9\ddagger}$} \quad \textbf{Sirui Li$^{10\ddagger}$} \\[4pt]
\textbf{Cathy Wu$^{1\dagger}$} \quad \textbf{Bryan Kian Hsiang Low$^{2,8\dagger}$} \quad \textbf{Jinhua Zhao$^{1,8\dagger}$} \quad \textbf{Paul Pu Liang$^{1,8\dagger}$} \\[8pt]
$^{1}$MIT \quad $^{2}$NUS \quad $^{3}$MiniMax \quad $^{4}$McGill \quad $^{5}$Stanford \quad $^{6}$SambaNova \quad $^{7}$Meta \\
$^{8}$Singapore-MIT Alliance for Research and Technology \quad $^{9}$Amazon \quad $^{10}$Microsoft \\[4pt]
{\small $*$ Equal contribution (alphabetical order) \quad $\dagger$ Joint advising} \\[4pt]
{\small $\ddagger$ Work done outside the authors' employment} \\[2pt]
}

\begin{document}
\begingroup
\renewcommand{\thefootnote}{}
\footnotetext{Correspondence: \texttt{qua@mit.edu}, \texttt{hanzheng@mit.edu}, \texttt{zhou\_zijian@u.nus.edu}}
\endgroup

\ifcolmsubmission
\linenumbers
\fi

\maketitle

\begin{center}
    \includegraphics[width=\textwidth]{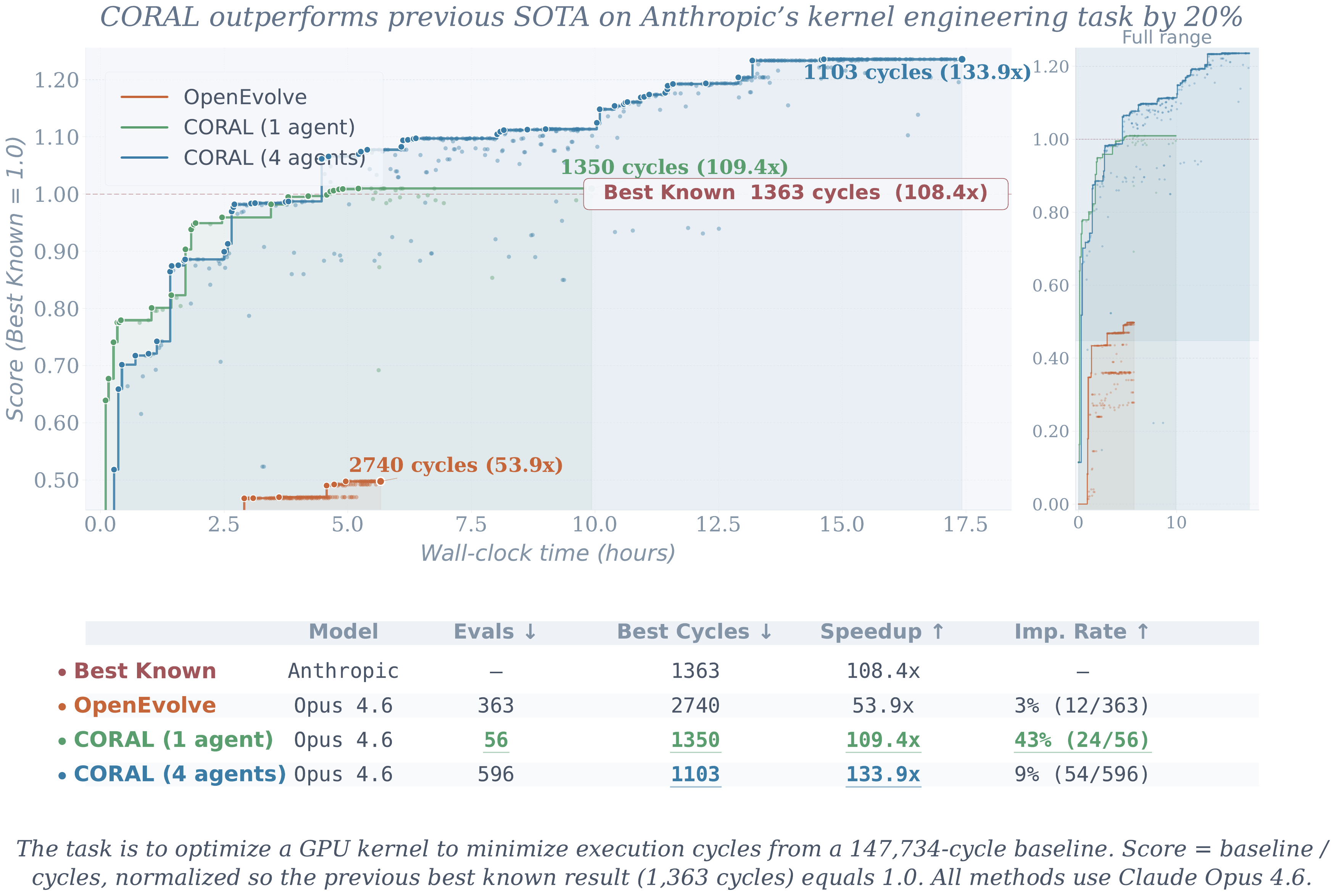}
    \label{fig:teaser}
\end{center}


\begin{abstract}

Large language model (LLM)-based evolution is a promising approach for open-ended discovery, where progress requires sustained search and knowledge accumulation. Existing methods still rely heavily on fixed heuristics and hard-coded exploration rules, which limit the autonomy of LLM agents. We present \textbf{CORAL}, the first framework for autonomous multi-agent evolution on open-ended problems. CORAL replaces rigid control with long-running agents that explore, reflect, and collaborate through shared persistent memory, asynchronous multi-agent execution, and heartbeat-based interventions. It also provides practical safeguards, including isolated workspaces, evaluator separation, resource management, and agent session and health management. Evaluated on diverse mathematical, algorithmic, and systems optimization tasks, CORAL sets new state-of-the-art results on 10 tasks, achieving 3--10$\times$ higher improvement rates with far fewer evaluations than fixed evolutionary search baselines across tasks. On Anthropic's kernel engineering task, four co-evolving agents improve the best known score from 1363 to 1103 cycles. Mechanistic analyses further show how these gains arise from knowledge reuse and multi-agent exploration and communication. Together, these results suggest that greater agent autonomy and multi-agent evolution can substantially improve open-ended discovery. Code is available at \url{https://github.com/Human-Agent-Society/CORAL}.

\end{abstract}


\vspace{-6pt}
\section{Introduction}
\vspace{-2pt}


Many important scientific problems do not come with ground-truth answers~\citep{mang2025frontiercs}. What is the best heuristic for a logistics problem~\citep{chen2025heurigym, zheng2026learning}? How should one write the most efficient kernel~\citep{ouyangkernelbench}? In these settings, the objective is clear, but the optimal solution is unknown. As a result, one-shot generation is insufficient. Strong solutions must be discovered through iterative proposal, testing, revision, and progress over time. 

Recent advances in LLM-powered agents have made this paradigm increasingly effective. Systems such as FunSearch~\citep{romera2024funsearch}, AlphaEvolve~\citep{novikov2025alphaevolve}, and more~\citep{lange2025shinkaevolve,agrawal2025gepa,cemri2026adaevolve} show that LLMs can be embedded in evaluator-guided evolutionary search loops for open-ended discovery. Rather than attempting to solve the problem in a single pass, these methods place the LLM inside an outer-loop search procedure: the model proposes candidate programs conditioned on previously high-scoring solutions, external evaluators execute and score these candidates under task-specific objectives, and a predetermined evolutionary algorithm governs parent selection and population updates. This type of LLM-based evolution has proven effective across mathematical discovery, algorithm design, and systems optimization tasks.

\begin{figure*}[h]
  \centering
  \includegraphics[width=\linewidth]{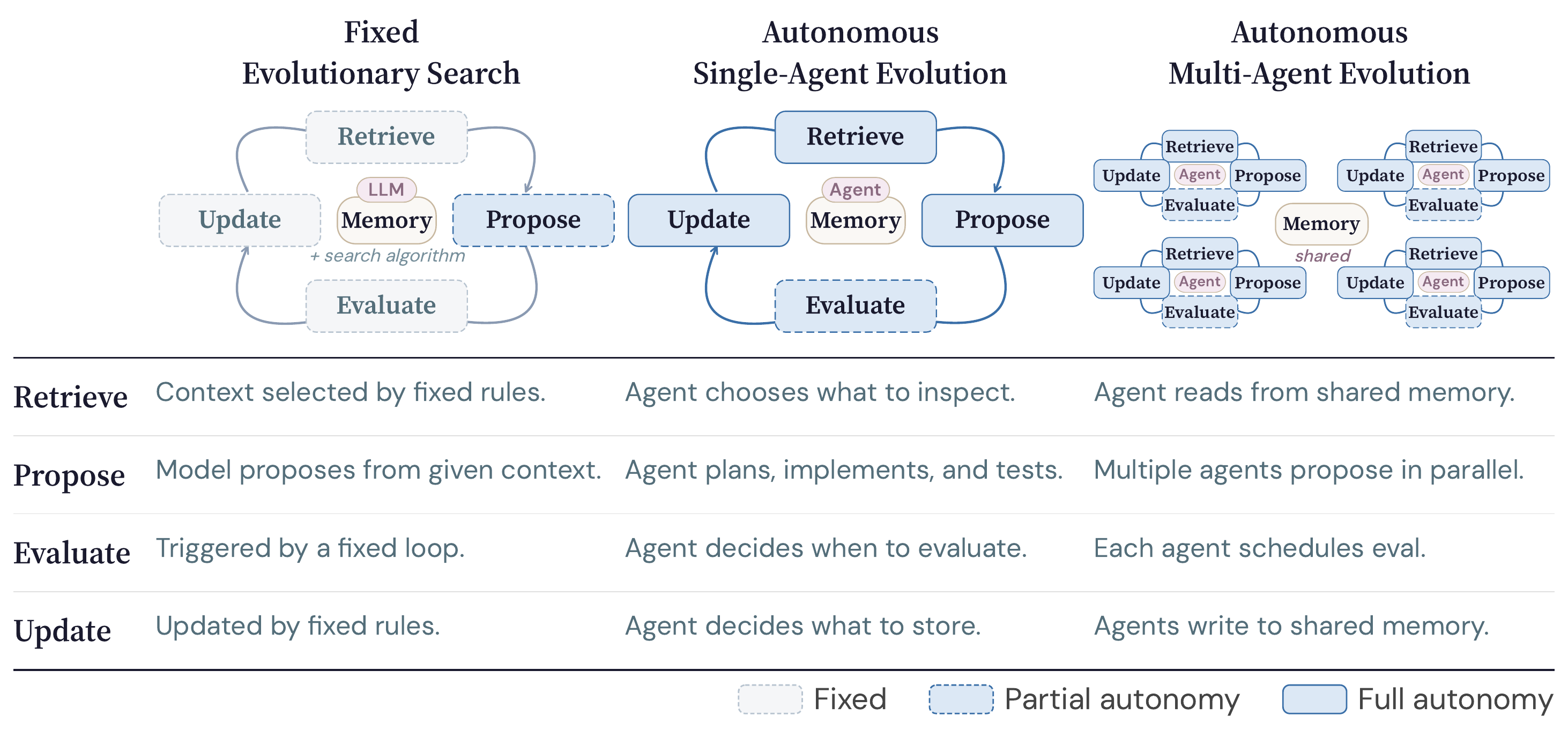}
  \caption{Comparison of three paradigms for LLM-based open-ended discovery.}
  \label{fig:paradigm_comparison}
\end{figure*}

However, current progress has largely been achieved with fixed evolutionary search, where key search decisions are independent of the agents, including which parent solutions to inspect and build on, when to run intermediate tests, and what knowledge to externalize and reuse later. For challenging open-ended problems, these choices are integral to the evolutionary algorithm and can substantially affect performance. This naturally leads to the following question: \textbf{Can stronger performance emerge when a greater part of the evolutionary algorithm is delegated to autonomous agents?}

The rigidity of these strategies is further amplified in multi-agent systems. Much of the existing literature relies on \emph{vertical scaling}: humans decompose the task, assign specialized roles, and define a fixed communication structure~\citep{hong2024metagpt,li2023camel,wu2024autogen}. While this paradigm has produced strong systems such as Sakana AI's AI Scientist~\citep{lu2024aiscientist} and Google's AI Co-Scientist~\citep{gottweis2025towards}, it assumes that the optimal decomposition and interaction topology are known in advance. For open-ended problems, that assumption is restrictive. This raises a further question: \textbf{Can multiple autonomous agents scale more effectively through \emph{horizontal parallelism}, by exploring in parallel, exchanging discoveries, and building on each other's progress over time?} Figure~\ref{fig:paradigm_comparison} illustrates the progression across these three paradigms.


As a key step towards this new paradigm, we introduce \method, a framework for autonomous multi-agent evolution on open-ended problems. \method\ shifts decision-making from fixed algorithms to the agents themselves, supported by a shared persistent memory for continuous evolution. Agents iteratively refine solutions by retrieving, contributing, and distilling knowledge in the form of notes and reusable skills into a collective repository. By incorporating heartbeat mechanisms for periodic reflection and redirection, \method\ ensures robust exploration and knowledge accumulation, providing a task-agnostic architecture compatible with diverse agent implementations.
Empirically, CORAL establishes new SOTA on $8$ of $11$ tasks in mathematical and systems optimization, with a $2.5\times$ higher improvement rate and $10\times$ fewer evaluations than fixed evolutionary search baselines. On the stress-test Kernel Engineering task, four co-evolving agents push the score from $1,363$ to $1,103$ cycles (a $20\%$ gain), surpassing the previous best result. Ablation studies confirm that both agent autonomy and multi-agent evolution contribute to these gains.

Our contributions are threefold. First, we formulate autonomous evolution as a distinct paradigm for open-ended discovery and distinguish autonomous single-agent and multi-agent evolution from prior fixed evolutionary search. Second, we introduce CORAL, a framework that realizes this paradigm through shared persistent memory, asynchronous multi-agent organization, and heartbeat-based interventions for long-horizon search. Third, across mathematical, algorithmic, and systems optimization tasks, we show that CORAL substantially outperforms fixed evolutionary search baselines, and ablations and trajectory analyses show the importance of knowledge accumulation and multi-agent evolution.

\section{Related Work}

\textbf{LLM-Driven Evolutionary Search.}
A growing line of work embeds LLMs as mutation operators within evaluator-guided evolutionary loops. FunSearch~\citep{romera2024funsearch} introduced this paradigm, and AlphaEvolve~\citep{novikov2025alphaevolve} extended it to full codebases with MAP-Elites. Subsequent systems refine the search orchestration through adaptive sampling, island-based architectures, and Pareto-based selection~\citep{sharma2025openevolve,lange2025shinkaevolve,khrulkov2025gigaevo,assumpcao2026codeevolve,yan2026pacevolve,agrawal2025gepa}, while AdaEvolve~\citep{cemri2026adaevolve} and EvoX~\citep{liu2026evox} make the search strategy itself adaptive. A complementary direction fine-tunes the generator at test time~\citep{wang2025thetaevolve,yuksekgonul2026tttdiscover} or builds experience libraries from solver feedback~\citep{ouyang2025reasoningbank,kong2025alphaopt}. All these systems follow a fixed pipeline: select parents via predefined heuristics, construct a prompt, and call the LLM to produce a mutation. The LLM has no agency over what to explore next. CORAL removes this scaffolding and lets the agent decide what to explore and what knowledge to carry forward.

\textbf{Autonomous LLM Agents.}
A separate line of work grants LLM agents the autonomy to carry out open-ended tasks without rigid external scaffolding. Autonomous coding agents~\citep{yang2024sweagent,wang2024openhands,openclaw2026,ding2026wildclawbenchbenchmarkrealworldlonghorizon} navigate codebases, execute code, and iteratively debug within sandboxed environments, while the AI Scientist~\citep{lu2024aiscientist} automates the full research cycle. Self-improvement techniques such as verbal self-feedback~\citep{shinn2023reflexion,madaan2023selfrefine}, interleaved reasoning and tool use~\citep{yao2023react}, and learned memory consolidation~\citep{zhou2026mem,yu2026memagent} further extend agent capabilities over long horizons. Recent position papers argue for elevating deployment-time adaptation to an autonomous evolver agent~\citep{gao2026agentic}. These systems demonstrate the power of agent autonomy, but they target one-off task completion rather than sustained, goal-driven optimisation. CORAL brings this autonomy into the evolutionary loop, replacing rigid search heuristics with agent-level intelligence at each evolution step.

\textbf{Multi-Agent Collaboration.}
Multi-agent LLM systems decompose complex tasks through role assignment and structured communication~\citep{wu2023autogen,langgraph2024,qian2024chatdev,hong2024metagpt}, or explore emergent cooperation via role-playing and dynamic group formation~\citep{li2023camel,chen2024agentverse}.
In existing evolutionary systems such as FunSearch and AlphaEvolve, parallelism is limited to running multiple stateless evaluation workers concurrently with no memory across steps.
CORAL introduces long-lived, stateful agents that communicate asynchronously through shared knowledge (scored attempts, notes, and skills), enabling emergent behaviours such as technique diffusion, spontaneous consensus, and cross-referencing, none of which are hardcoded.


\section{Coral: A Framework for Autonomous Multi-Agent Evolution}
\subsection{Preliminaries: Problem Formulation for Open-Ended Discovery}
\label{Sec:method:preliminary}

We consider open-ended discovery tasks, where the optimal solution is unknown and increasingly strong candidate solutions must be discovered through iterative search under evaluator feedback. A task instance is specified by a task description $x$ and an evaluator $E$, where evaluating a candidate solution $y$ returns $E(x,y) := (s,f),$ with $s$ denoting the score of $y$ and $f$ denoting auxiliary feedback such as sub-score breakdowns or textual critique from an LLM-powered evaluator.

Let $\mathcal{M}_t$ denote the shared persistent memory available at search step $t$, such as prior candidate solutions and their evaluation outcomes. At an abstract level, each improvement step consists of four stages:
\squishlisttwo[\arabic*.]
    \item \stage{Retrieve}: construct a working context $\hat{\mathcal{M}}_t$ from $\mathcal{M}_t$;
    \item \stage{Propose}: generate a candidate solution $y_{t+1}$ conditioned on $x$ and $\hat{\mathcal{M}}_t$;
    \item \stage{Evaluate}: obtain score and feedback $(s_{t+1}, f_{t+1}) = E(x, y_{t+1})$;
    \item \stage{Update}: incorporate new information into shared persistent memory to form $\mathcal{M}_{t+1}$.
\squishend

\subsection{From Fixed Search to Autonomous Multi-Agent Evolution}






Most prior LLM-based methods for open-ended discovery follow \emph{fixed evolutionary search}, where the four stages in Section~\ref{Sec:method:preliminary} are instantiated by externally specified rules (Figure~\ref{fig:paradigm_comparison}). In this paradigm, \stage{Retrieve} and \stage{Update} are governed by fixed procedures, while the LLM mainly acts in \stage{Propose}, typically generating a candidate from a constructed context in a single forward pass, and \stage{Evaluate} is handled by the task evaluator. For example, in AlphaEvolve~\citep{novikov2025alphaevolve}, the working context $\hat{\mathcal{M}}_t$ is constructed from $\mathcal{M}_t$ using predetermined selection rules inspired by MAP-Elites and island models. This paradigm is effective, but it leaves key search decisions outside the agent. The agent does not decide what evidence to inspect, when to verify intermediate results, how to react to failure, or what knowledge to preserve for reuse. For open-ended discovery, however, these choices are often part of the problem itself.

This motivates \textbf{\emph{autonomous single-agent evolution}}. Here, a single agent controls a much larger portion of the search process: it can decide what to retrieve, when to run local tests, when to invoke the evaluator, and what to write back to persistent memory. The same four stages still apply, but their timing and realization are no longer fixed externally (Figure~\ref{fig:paradigm_comparison}). We further extend this idea to \textbf{\emph{autonomous multi-agent evolution}}, where multiple agents run asynchronously while coordinating through shared persistent memory (Figure~\ref{fig:paradigm_comparison}). Rather than relying on predefined roles or a communication structure, agents interact indirectly through shared persistent memory. This increases exploration diversity and allows multiple agents to inspire each other.

We advocate autonomous multi-agent evolution as a promising paradigm for open-ended discovery and introduce \method{} as a lightweight infrastructure to realize it. \method{} delegates much more of the search process to autonomous agents, while keeping the evaluator as an API accessible to the agent, with the grader details hidden.
This added flexibility also introduces systems challenges: agents must remain persistent over long horizons, avoid drift, accumulate reusable knowledge, and operate safely without overloading compute resources or hacking the evaluator. To address these challenges, \method{} introduces three core mechanisms: shared persistent memory, asynchronous multi-agent organization, and heartbeat-based interventions, along with several execution safeguards (see Appendix~\ref{app:safeguards}).

\begin{figure*}[t!]
  \centering
  \includegraphics[width=\linewidth]{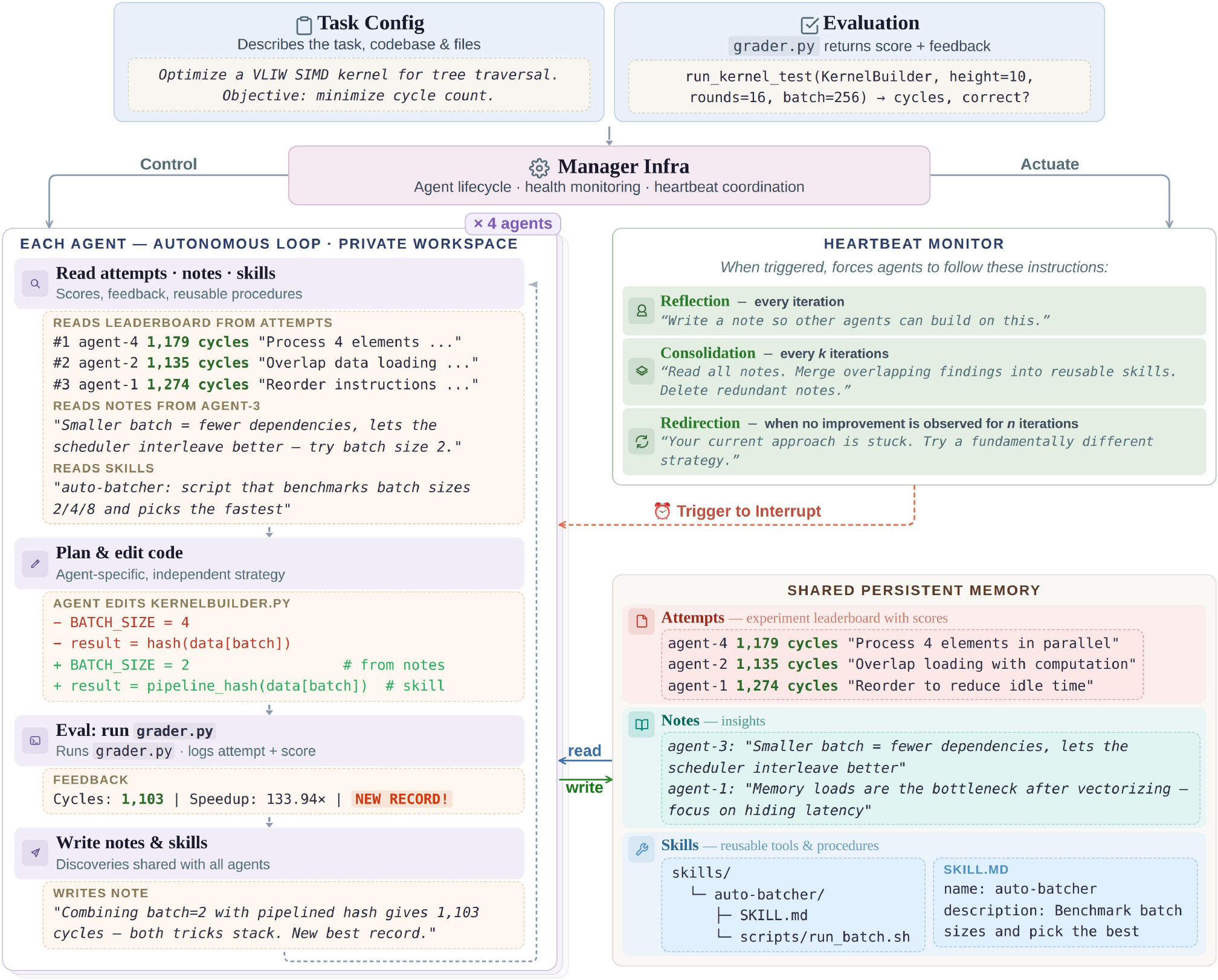}
  \vspace{-1pt}
  \caption{Overview of the \method{} framework. Autonomous agents operate in isolated worktrees, iteratively propose and evaluate candidate solutions, and accumulate shared persistent memory (attempts, notes, skills) through a hub. Heartbeat-driven periodic reflections help agents consolidate discoveries and reorient search over long horizons.}
  \label{fig:coral-diagram}
\end{figure*}

\subsection{Core Mechanisms of \method{}}

\textbf{Shared Persistent Memory as File System.}
\method{}'s shared persistent memory $\mathcal{M}$ is structured as a file system with symbolic links to an agent's workspace (also a file system) to maintain consistency. Functioning much like a library, an agent can retrieve and contribute to the shared file system via the \method{} CLI tool (see Appendix~\ref{app:apis}) and directly use \texttt{Bash} tool to access it. Agents can even help `organize` the shared persistent memory by categorizing shared knowledge into sub-folders. This design allows progressive disclosure for agents, saving their contexts, while also being easy to maintain and highly extensible. To provide some bootstrapping structure for the agents, we define three root folders storing different types of knowledge, explained below. Please refer to Appendix~\ref{app:shared_memory} for examples.



\textbf{\texttt{attempts/}} records historical evaluations and solutions. Agents can browse this space to understand high-performing solutions and retrieve their solutions for comparison.

\textbf{\texttt{notes/}} records observations, learnings, and reflections from all agents. Each note is stored as a markdown file in the directory or a subdirectory determined by the agent who creates the note. Agents have full access to all notes.

\textbf{\texttt{skills/}} records reusable procedures, tools, scripts, and implementation patterns transferable across attempts. Following standard practice, a skill consists of a natural-language description (e.g., \texttt{SKILL.md}) together with executable artifacts such as functions and example scripts. Agents are provided with a \texttt{skill\_creator} skill as a guide to create new skills.

\textbf{Multi-Agent Organization.}
\method{} naturally extends from a single autonomous agent to a population of \(N\) agents that run asynchronously. Each agent \(i\) maintains its own local context \(\mathcal{C}_t^{(i)}\) and executes in an isolated workspace while sharing access to the same evaluator and shared persistent memory \(\mathcal{M}\) via symbolic link, i.e., shortcut pointers to the original files. This design allows each agent to freely work on its own without interference.


Unlike many peer-to-peer multi-agent systems where agents directly talk to each other~\citep{langgraph2024}, coordination between agents occurs primarily through shared persistent memory. Similar to the single-agent scenario, each agent may autonomously read and write to a shared workspace. As agents generate attempts, notes, and skills, they write artifacts \(\mathcal{W}_t^{(i)}\) to \(\mathcal{M}\), which may later be retrieved by other agents as part of their own context \(\hat{\mathcal{M}}_t^{(j)}\). 
This way, one agent's discoveries can influence another agent's future search through what it writes to the shared workspace, without requiring a messaging protocol. 
This organization increases exploration diversity by allowing agents to pursue different local directions in parallel while still benefiting from shared accumulation.

\textbf{Heartbeat: Reflection, Consolidation and Redirection.}
As \method{} does not enforce the fixed-search workflow, agents may inadvertently fall into local minima, where they decide to work on micro-optimizations instead of trying out innovative ideas. Agents may also forget to consult and contribute to shared persistent memory. To encourage desirable behavior, \method{} imposes a heartbeat mechanism that functions like a \textit{Reminder App}, periodically prompting the agents to exercise self-reflection and pivoting for new ideas when existing approaches plateau.
A heartbeat event may be predefined or created by the agents themselves. Each event is attached to a trigger, which can be the number of intervals, time passed, or a change in eval score.
When triggered at step $t$, the heartbeat applies a modification to the agent's local context $\mathcal{C}_t^{(i)} \rightarrow \mathcal{C}_t^{\prime (i)}$ and thereby steers subsequent behavior.

\method{} implements three heartbeat types. The first is a \emph{per-iteration reflection heartbeat}, which encourages the agent to record useful notes during ongoing work. This heartbeat helps the agent capture observations as they arise. The second is a \emph{periodic consolidation heartbeat}, triggered after a fixed number of attempts, which prompts the agent to review progress, organize and refine accumulated notes, and distill reusable procedures into skills. In other words, while the first supports note-taking during work, the second focuses on organizing those notes and building skills from them. The third is a \emph{stagnation-triggered redirection heartbeat}, activated when the agent shows no improvement for several rounds, which prompts it to reassess the current direction and decide whether to continue, revise its strategy, or pivot to a different line of search. Together, these heartbeat mechanisms promote explicit memory formation and reduce myopic local search.



\vspace{-4pt}
\section{Experiments}
\vspace{-4pt}

\subsection{Experimental Setup}
\label{sec:exp_setup}

\textbf{Tasks.}
We evaluate CORAL on two benchmark suites and two stress-test problems.
The benchmark suites follow the experiment set-up in EvoX~\citep{liu2026evox} and TTT-Discover ~\citep{yuksekgonul2026tttdiscover}, consisting of 6 mathematical optimization tasks (e.g., circle packing, Erd\H{o}s minimum overlap) and 5 systems optimization tasks (e.g., expert placement load balancing, GPU placement, cross-cloud transfer). These suites are used for both single-agent and multi-agent experiments.
The two challenging stress-test problems are further used for multi-agent evaluation: Anthropic's kernel engineering ~\citep{anthropic_takehome}, a VLIW SIMD tree-traversal task with an official best score of 1,363 cycles, and the Polyominoes packing problem from Frontier-CS~\citep{mang2025frontiercs}, which is one of the hardest among all 172 problems in the benchmark.

\textbf{Baselines and models.}
For single-agent experiments, we compare CORAL against OpenEvolve~\citep{sharma2025openevolve}, ShinkaEvolve~\citep{lange2025shinkaevolve}, and EvoX~\citep{liu2026evox}, all given the same seed programs, evaluators, and budgets. All single-agent methods and the stress-test multi-agent experiments use Claude Code + Opus 4.6. For multi-agent experiments on the math and systems suites, we use a fully open-source stack (MiniMax M2.5~\citep{minimax2026m25} + OpenCode~\citep{opencode2025}) to verify that CORAL's gains generalize beyond proprietary models and agents. No internet access is provided.

\textbf{Budget and evaluation.}
All runs on the math and systems suites are given a 3-hour wall-clock budget or 100 iterations for the baseline methods, whichever is longer. For fairness, we run CORAL for the minimum duration among all baseline runs. Stress-test problems run until convergence due to difficulty. All results are averaged over 4 independent trials. We report:
\squishlisttwo
  \item \textbf{Final score}: the best score achieved within the evaluation budget (primary metric).
  \item \textbf{Improvement rate}: fraction of evaluations that yield an improvement on the best score.
  \item \textbf{Number of evaluations}: number of evaluations required to reach the final score.
\squishend
{\setlength{\fboxsep}{1pt}\colorbox{myyellow!15}{SOTA}} denotes the best previously known results (human or AI).

\subsection{Autonomous Evolution Outperforms Fixed Evolutionary Search}
\label{sec:exp_main_single}

As shown in Table~\ref{tab:main_results}, \method{} \textbf{achieves the best final score on all 11 tasks, establishing new SOTA on 8 tasks}. Among baselines, EvoX~\citep{liu2026evox} is the strongest competitor due to its meta-evolved search strategy, yet CORAL still outperforms it on every task. All three baselines trail CORAL by a wide margin on improvement rate and evaluation efficiency. CORAL's improvement rate is 3 -- 10$\times$ higher, and it typically converges within 5 -- 20 evaluations versus 60 -- 100 for fixed evolutionary search methods. This means CORAL wastes far fewer evaluations on unproductive candidates. We attribute this to CORAL's autonomous design. fixed evolutionary search baselines select candidates for mutation based on predefined heuristics and follow a fixed pipeline at each evolution step. CORAL agents instead decide what to explore next based on their own analysis of prior attempts and evaluation feedback, choosing which aspects of a solution to modify, when to pivot to a different approach, and when a candidate is ready for evaluation. This autonomy over the evolutionary process is reflected directly in the gap in improvement rates.

\renewcommand{\arraystretch}{0.96}
\begin{table}[t]
\caption{Single-agent \method{} vs.\ fixed evolutionary search baselines on mathematical and systems optimization tasks. OE = OpenEvolve, SE = ShinkaEvolve. All methods use Claude Opus 4.6. For Final Score, $\uparrow$ means higher is better and $\downarrow$ means lower is better. For Improvement Rate, higher is better. For \# Evals, lower is better. {\setlength{\fboxsep}{1pt}\colorbox{cyan!10}{Cyan}} cells surpass previous {\setlength{\fboxsep}{1pt}\colorbox{myyellow!15}{SOTA}} on final score. Best results are \textbf{bolded}. CORAL's autonomous evolution significantly outperforms fixed evolutionary search, achieving the best final score on all 11 tasks and establishing new SOTA on 8 tasks.}
\label{tab:main_results}
\centering
\scriptsize
\setlength{\tabcolsep}{2.7pt}
\vspace{-7pt}
\begin{threeparttable}
\begin{tabular}{c c | >{\columncolor{myyellow!15}}c cccc | cccc | cccc}
\toprule
& \multirow{3}{*}{\textbf{Task}} & \multicolumn{5}{c|}{\textbf{Final Score}} & \multicolumn{4}{c|}{\textbf{Impr.\ Rate (\%)}} & \multicolumn{4}{c}{\textbf{\# Evals}} \\ \cmidrule(lr){3-7}  \cmidrule(lr){8-11} \cmidrule(lr){12-15}
& & \textbf{SOTA} & \textbf{OE} & \textbf{SE} & \textbf{EvoX} & \textbf{CORAL} & \textbf{OE} & \textbf{SE} & \textbf{EvoX} & \textbf{CORAL} & \textbf{OE} & \textbf{SE} & \textbf{EvoX} & \textbf{CORAL} \\
\midrule
\multirow{6}{*}{\rotatebox{90}{\textbf{Math}}}
& Circle-Pack.\,$\uparrow$         & 2.6359 & 2.6293 & 2.6001 & 2.6320 & \hl{\textbf{2.6360}} & 7.0 & 19.4 & 27.1 & \textbf{100.0}   & 100 & 62  & 48  & \textbf{11} \\
& Signal Proc.\,$\uparrow$         & 0.7429 & 0.6420 & \hl{0.8171} & 0.7306 & \hl{\textbf{0.8229}} & 11.8 & 8.6 & 21.1 & \textbf{30.3} & 85  & 58  & 71  & \textbf{56} \\
& Erd\H{o}s Over.\,$\downarrow$    & \textbf{0.38088} & 0.38188 & 0.38156 & 0.38125 & 0.38089 & 9.5 & 15.7 & 27.8 & \textbf{36.8} & 84  & 89  & 36  & \textbf{19} \\
& MMD-16-2\,$\downarrow$           & \textbf{12.89} & 12.92 & \textbf{12.89} & 12.96 & \textbf{12.89} & 8.2 & 9.8 & 33.3 & \textbf{83.3} & 97  & 82  & 18  & \textbf{6}  \\
& MMD-14-3\,$\downarrow$           & \textbf{4.16} & 4.21 & 4.46 & 4.46 & \textbf{4.16} & 25.0 & 9.4 & 24.5 & \textbf{75.0} & 12  & 96  & 53  & \textbf{8}  \\
& 3rd-Autocorr.\,$\downarrow$      & \textbf{1.4557} & 1.4731 & 1.4812 & 1.5552 & \textbf{1.4557} & 6.2 & 32.4 & 12.0 & \textbf{60.0} & 97  & 37  & 75  & \textbf{5}  \\
\midrule
\multirow{5}{*}{\rotatebox{90}{\textbf{System}}}
& EPLB\,$\uparrow$                 & 0.145 & 0.127 & 0.129 & \hl{0.146} & \hl{\textbf{0.149}} & 8.3 & 12.6 & 10.0 & \textbf{78.9} & 60  & 87  & 100 & \textbf{19} \\
& PRISM\,$\uparrow$                & \textbf{26.26} & \textbf{26.26} & \textbf{26.26} & \textbf{26.26} & \textbf{26.26} & 31.6 & 18.8 & 25.0 & \textbf{100.0}   & 19  & 32  & 16  & \textbf{3}  \\
& LLM-SQL\,$\uparrow$              & 0.730 & 0.716 & 0.724 & 0.726 & \hl{\textbf{0.731}} & 6.7 & 23.8   & 6.0 & \textbf{53.3} & 100 & 21  & 83  & \textbf{15} \\
& Txn Sched.\,$\uparrow$           & 4348 & 3774 & 3802 & 3984 & \hl{\textbf{4566}} & 10.1 & 16.0 & 16.0 & \textbf{27.3} & 99  & 25  & 94  & \textbf{22}  \\
& Cloudcast\,$\downarrow$          & 632.7 & \hl{627.2} & \hl{627.8} & \hl{623.5} & \hl{\textbf{618.4}} & 7.2 & 20.8 & 12.3 & \textbf{33.3} & 69  & 24  & 65  & \textbf{9} \\
\bottomrule
\end{tabular}
\end{threeparttable}
\end{table}

\subsection{Multi-Agent Evolution Extends the Search Frontier}
\label{sec:exp_main_multi}

\textbf{Multi-agent gains over strong single-agent autonomy.}
While single-agent \method{} already outperforms all fixed evolutionary search baselines, 4-agent co-evolution pushes performance even further (Table~\ref{tab:multi-agent}). The largest improvements appear on the stress-test problems, where single-agent runs tend to plateau early, with co-evolution achieving an 18.3\% cycle reduction on Kernel Engineering and a 5.0\% score increase on Polyominoes. Notably, without web search, CORAL already \textbf{establishes a new SOTA on the Kernel Engineering task}. With web search enabled, CORAL also \textbf{achieves a new SOTA (89.4) on the Polyominoes problem}, although for fairness, we report the non-web-search results in Table~\ref{tab:multi-agent}; the full web-enabled results are provided in Appendix~\ref{app:polyominoes_sota}. These gains do not arise solely from additional compute: single-agent runs exhibit higher per-eval improvement rates, yet co-evolution achieves better final scores by exploring more diverse search trajectories. This is enabled by CORAL's asynchronous shared persistent memory, where multiple agents independently explore different regions of the solution space and share discoveries through persistent attempts, notes, and skills. Useful techniques diffuse across agents organically without requiring explicit coordination protocols. 

\renewcommand{\arraystretch}{0.98}
\begin{table}[t]
\caption{Multi-agent co-evolution vs.\ single-agent CORAL. \textbf{Bolded} Gain (\%) values denote the relative improvement of 4-Agents over 1-Agent. Multi-Agent evolution can significantly improve the search frontier, especially on tasks where single-agent runs plateau early.}
\label{tab:multi-agent}
\centering
\scriptsize
\setlength{\tabcolsep}{4pt}
\begin{threeparttable}
\begin{tabular}{c c c | >{\columncolor{myyellow!15}}c ccc | cc | cc}
\toprule
& \multirow{3}{*}{\textbf{Runtime}} & \multirow{3}{*}{\textbf{Task}} & \multicolumn{4}{c|}{\textbf{Final Score}} & \multicolumn{2}{c|}{\textbf{Impr.\ Rate (\%)}} & \multicolumn{2}{c}{\textbf{\# Evals}} \\  \cmidrule(lr){4-7}  \cmidrule(lr){8-9} \cmidrule(lr){10-11}
&   &  & \textbf{SOTA} & \textbf{1-Agent} & \textbf{4-Agent} & \textbf{Gain (\%)} & \textbf{1-Agent} & \textbf{4-Agent} & \textbf{1-Agent} & \textbf{4-Agent} \\
\midrule
& \multirow{2}{*}{\shortstack{Claude Code +\\Opus 4.6}}
& Polynominoes\,$\uparrow$        & 87.0 & 80.2   & 84.2   & \textbf{4.99}  &  42.4     &   19.4    &  33   &  67   \\[-1pt]
\multirow{-2}{*}{\rotatebox{90}{\textbf{Stress}}}
& & Kernel Eng.\,$\downarrow$          & 1363 & 1350   & 1103   & \textbf{18.30} &    43.0   &  9.0     &  56   &  596   \\[2pt]
\midrule
\multirow{6}{*}{\rotatebox{90}{\textbf{Math}}}
& \multirow{6}{*}{\shortstack{OpenCode +\\MiniMax M2.5}}
& Circle-Pack.\,$\uparrow$         & 2.6359 & 2.3531 & 2.5391 & \textbf{7.90}  & 20.7 & 28.3 & 29  & 46  \\
& & Signal Proc.\,$\uparrow$         & 0.7429 & 0.7174 & 0.7383 & \textbf{2.91}  & 30.5 & 19.8 & 59  & 253 \\
& & Erd\H{o}s Over.\,$\downarrow$    & 0.38088 & 0.39237 & 0.38311 & \textbf{2.36} & 16.7 & 6.9 & 42  & 58  \\
& & MMD-16-2\,$\downarrow$           & 12.89 & 12.91  & 12.89  & \textbf{0.15}  & 32.4 & 11.5 & 34  & 103 \\
& & MMD-14-3\,$\downarrow$           & 4.16 & 4.53   & 4.19   & \textbf{7.51}  & 63.6 & 15.0 & 11  & 80  \\
& & 3rd-Autocorr\,$\downarrow$       & 1.4557 & 1.5337 & 1.4931 & \textbf{2.65}  & 42.9 & 22.1 & 7   & 59  \\
\midrule
\multirow{5}{*}{\rotatebox{90}{\textbf{System}}}
& \multirow{5}{*}{\shortstack{OpenCode +\\MiniMax M2.5}}
& EPLB\,$\uparrow$                 & 0.145 & 0.128  & 0.129  & \textbf{0.78}  & 50.0   & 65.4 & 6   & 26  \\
& & PRISM\,$\uparrow$                & 26.26 & 25.85  & 26.26  & \textbf{1.59}  & 32.6 & 31.7 & 46  & 82  \\
& & LLM-SQL\,$\uparrow$              & 0.730 & 0.693  & 0.730  & \textbf{5.34}  & 83.3 & 71.7 & 6   & 46  \\
& & Txn Sched.\,$\uparrow$           & 4348 & 3704   & 3774   & \textbf{1.89}  & 50.0   & 66.7 & 16  & 24  \\
& & Cloudcast\,$\downarrow$          & 632.7 & 849.4  & 672.8  & \textbf{20.80} & 72.7 & 66.7 & 11  & 12  \\
\bottomrule
\end{tabular}
\end{threeparttable}
\end{table}

\textbf{Generalization to open-source models.}
The multi-agent gains are not tied to proprietary models. When evaluated on the math and systems suites using a fully open-source stack (MiniMax M2.5 + OpenCode), 4-agent co-evolution consistently improves final scores over the single-agent counterpart across most tasks (Table~\ref{tab:multi-agent}). These results show that CORAL's organizational advantages arise from the co-evolution mechanism itself rather than model-specific capabilities, and that the benefits of distributed exploration and shared persistent memory transfer to open-source settings.

\vspace{-4pt}
\subsection{Analysis}
\label{sec:exp_analysis}
\vspace{-4pt}

\subsubsection{Why Autonomous Evolution Works}
To understand why autonomous evolution is effective, we analyze agent trajectories qualitatively and quantitatively. Detailed results are reported in Appendix~\ref{app:trajectory_statistics}. Across tasks, local verification and knowledge accumulation are strongly associated with performance: 

\textbf{Local verification.} Agents often execute code and run tests locally before submitting for external evaluation, allowing them to debug and validate candidates within a single iteration. Attempts with local execution improve more often than the average attempt on the same task. This effect is the strongest on tasks involving compiled code: on Transaction (61\% local test rate) and Kernel Engineering (57\%) (see Table~\ref{tab:single_agent_trajectory_analysis} and~\ref{tab:task_level_trajectory_analysis}), local execution often catches compilation failures before an evaluation is consumed. By contrast, tasks with non-reproducible or hidden evaluations show much lower local test rates; for example, Prism (0\%) relies on grader-generated randomized tests.

\textbf{Knowledge accumulation.} Knowledge accumulation through notes and skills also helps, but its role differs sharply across task types. On standard tasks, agents create only 0.05 knowledge artifacts per attempt, and knowledge access yields only a small gain (+2 percentage points over attempts without knowledge access). On advanced tasks, agents create over 10$\times$ more knowledge per attempt (0.55 and 0.68), and knowledge access is much more strongly associated with improvement: 55\% on Kernel Engineering versus 26\% on standard tasks (see Table~\ref{tab:single_agent_trajectory_analysis}). The knowledge itself also differs in quality. On standard tasks, notes are often lightweight progress logs, such as records of parameter changes. On advanced tasks, they capture reusable insights: for example, Kernel Engineering notes identify architectural bottlenecks such as VALU or record cases where relaxing WAR dependencies hurts performance, while Polyominoes includes a ``what NEVER worked'' folder to document failed approaches and avoid revisiting unpromising design strategies across attempts.

Agents also proactively inspect prior attempts, compare implementations, and look for patterns when deciding what to try next. However, whether this form of inspection is more effective than retrieval in earlier fixed evolutionary search methods is difficult to isolate, so we leave it to future work.





\subsubsection{Why Multi-Agent Organization Helps}
\label{sec:exp_analysis_multi}
We analyze 4-agent runs on Kernel Engineering (596 attempts) and Polyominoes (67 attempts) across three dimensions.

\textbf{Cross-agent information transfer.}
Building on another agent's work is very effective. On Kernel Engineering, 36\% of attempts use another agent's commit as their parent, and these improve at 17\% versus 9\% for all attempts. The majority (66\%) of new records originate from a cross-agent parent. On Polyominoes, direct code transfer is rarer (12\%) but still very powerful (50\% versus a 19\% average improvement rate); transfer instead occurs more often through shared notes and skills, with 87\% of rounds referencing knowledge committed by other agents. The two tasks exhibit complementary information transfer modes: Kernel Engineering agents transfer more through referencing others' code, whereas Polyominoes agents transfer more through knowledge.

\textbf{Exploration diversity.}
We extract strategy keywords from attempt titles and compute pairwise Jaccard similarity. On Kernel Engineering, agents average 0.43 pairwise overlap; on Polyominoes, 0.31. More than half of each agent's strategy vocabulary is unique, meaning that the population collectively explores substantially more of the search space than any individual agent.

\renewcommand{\arraystretch}{0.97}
\begin{wraptable}{r}{0.34\textwidth}
\vspace{-20pt}
\caption{Ablation study on knowledge accumulation and multi-agent co-evolution. All runs use Claude Code + Opus 4.6. Best results are \textbf{bolded}.}
\label{tab:ablations}
\centering
\scriptsize
\setlength{\tabcolsep}{3pt}

\begin{tabular}{lcc}
\toprule

\multicolumn{3}{c}{\textbf{Knowledge Accumulation (1-Agent)}} \\

\textbf{Task} & \textbf{w/ Know.} & \textbf{w/o Know.} \\
\midrule
Kernel Eng.\,$\downarrow$  & \textbf{1350} & 1601 \\
Polyominoes\,$\uparrow$    & \textbf{80.2} & 77.3 \\
Txn Sched.\,$\uparrow$     & \textbf{4566} & 4444 \\

\midrule

\multicolumn{3}{c}{\textbf{Co-evolution (4-Agent)}} \\
\textbf{Task} & \textbf{Co-evol.} & \textbf{Indep.\ Best} \\
\midrule
Kernel Eng.\,$\downarrow$  & \textbf{1103} & 1180 \\
Polyominoes\,$\uparrow$    & \textbf{84.2} & 80.8 \\
Txn Sched.\,$\uparrow$     & \textbf{4694} & 4629 \\

\bottomrule
\end{tabular}
\vspace{-25pt}
\end{wraptable}

\textbf{Contribution balance.}
On Kernel Engineering, all four agents produce 130--165 attempts with 10--16 improvements each, and all four independently reach the best score of 1103 cycles. Records are evenly split (14/15/10/15). Leader tenure is more skewed: agent-1 holds the best score for 45\% of the run. On Polyominoes, contributions are less balanced: agent-3 sets 6 of 13 records, and agent-4 leads for 34\% of the total time.

\subsubsection{Ablations}
\label{sec:ablations}

We ablate two core components of \method{}: knowledge accumulation and multi-agent co-evolution. Table~\ref{tab:ablations} reports results on three stress-test tasks using Claude Code + Opus 4.6.

\textbf{Knowledge accumulation.} Disabling note and skill creation degrades final scores across all three tasks, with the largest drop on Kernel Engineering ($1350 \to 1601$ cycles, an 18.6\% regression). This confirms that knowledge artifacts causally contribute to search quality, rather than being merely correlated with improvement.

\textbf{Co-evolution vs.\ independent runs.} To test whether multi-agent evolution gains come from co-evolution or simply from running more agents, we compare 4-agent co-evolution against the best-of-4 independent single-agent runs. Co-evolution outperforms independent best on all three tasks. This shows that the gains from multi-agent evolution are not reducible to additional compute.


\section{Conclusion}

We introduced \method{}, a framework for autonomous multi-agent evolution for open-ended problems. By replacing rigid evolutionary search heuristics with autonomous agents that control retrieval, proposal, evaluation, and knowledge accumulation, while coordinating through shared persistent memory, \method{} achieves substantially stronger performance across mathematical, algorithmic, and systems optimization tasks. A single autonomous agent already outperforms fixed evolutionary-search baselines, and multi-agent evolution pushes the frontier further: four agents discover solutions that no single agent finds, even when the latter is given four times the compute. More broadly, our results suggest that autonomous agents are becoming a promising paradigm for open-ended discovery. Recent concurrent open-source projects~\citep{karpathy2026autoresearch,liu2026autoevolver,rllmorg2026hive} and emerging studies~\citep{chen2026avoagenticvariationoperators} point in a similar direction, and together with the empirical evidence in this work, they suggest that we may be approaching a turning point in how AI systems tackle problems that require iterative search, learning from feedback, and accumulation of knowledge over time. This progress is both exciting and unsettling, as it creates new opportunities for scientific and engineering discovery while also raising important challenges for the research community (Appendix~\ref{app:future_directions}). We hope \method{} can serve as a systematic exploratory study, a strong baseline framework, and an extensible infrastructure that supports future work on autonomous discovery systems.


\bibliography{colm2026_conference}
\bibliographystyle{colm2026_conference}

\clearpage


\appendix

\section*{LLM Usage Disclosure}
We used LLMs for minor writing assistance, including grammar correction and language polishing. The agents in CORAL are instantiated with LLMs, which serve as the backbone for all agent behaviors and experiments described in the paper. The core research ideas, methodology, experimental design, implementation, analysis, and conclusions were developed and carried out by the authors. No LLM was used to generate research ideas, experimental results, figures, or evaluations.

\section{Limitations and Future Directions}
\label{app:future_directions}
While \method{} has proved significantly effective across a wide range of challenging tasks, the current version still has several limitations. First, \method{} relies on frontier foundation models that can handle relatively complex coding-agent workflows, which makes full deployment on local devices difficult. An exciting direction for future work is therefore to train customized small models tailored to \method{}. Second, multi-agent evolution currently lacks bootstrapped heterogeneity: all agents are initialized identically and given access to the same information. Future work could inject distinct personalities, roles, or private information into different agents to encourage greater behavioral diversity and, in turn, a more efficient evolutionary process. Third, our current setting assumes the availability of a reasonably well-specified evaluator. However, for many important open-ended problems, evaluators are themselves difficult to obtain, incomplete, or even fundamentally ambiguous. In such settings, evaluation may also need to co-evolve with the solutions, for example through iterative refinement of the evaluator, learned critics, or human-agent negotiation over what constitutes progress. 
\section{Additional Experiment Results}
\label{app:experiment_results}

\subsection{New State-of-the-Art on Polyominoes Packing}
\label{app:polyominoes_sota}

\begin{figure*}[h]
  \centering
  \includegraphics[width=0.9\textwidth]{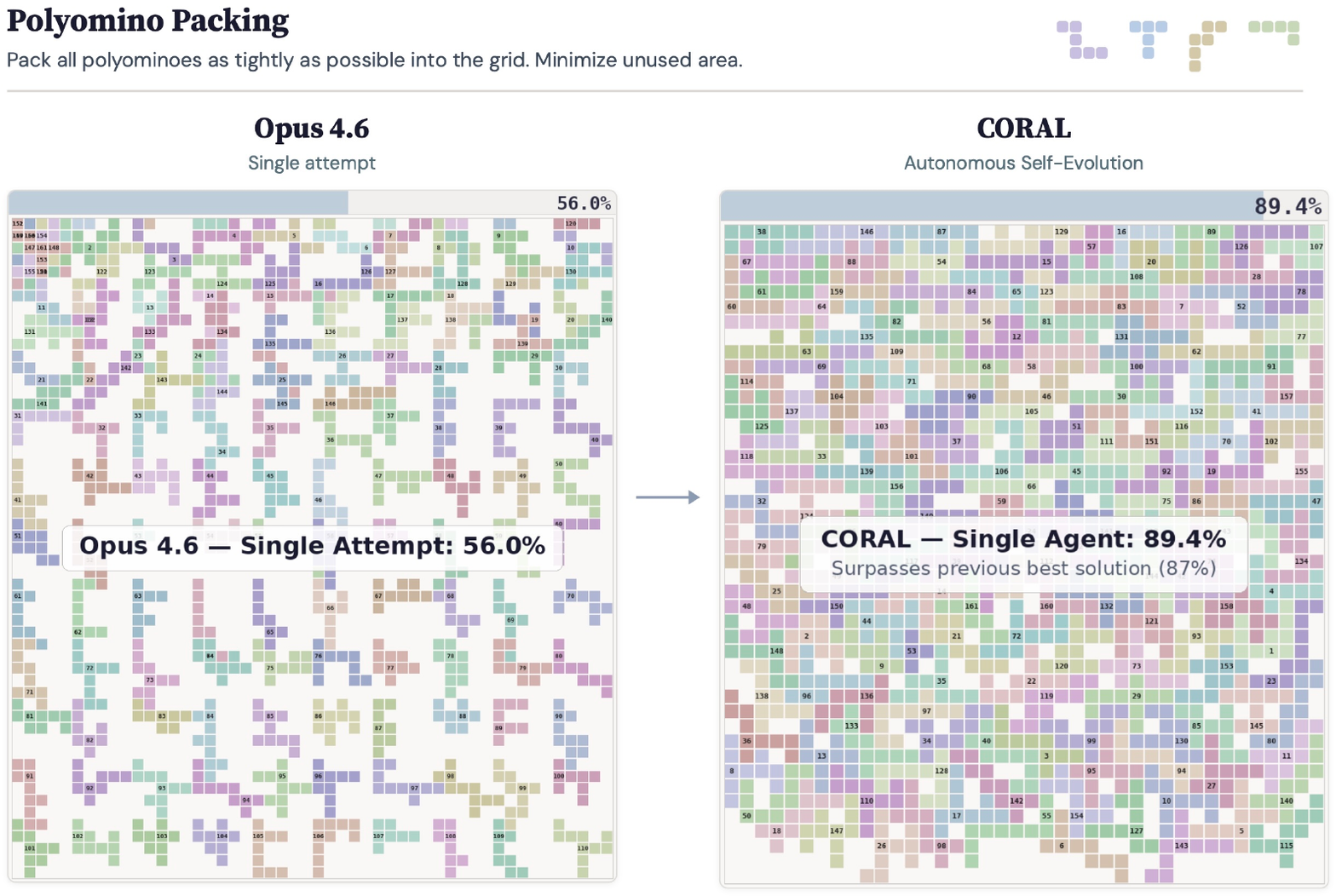}
  \caption{Polyominoes packing: single-attempt baseline (left, 56.0\%) vs.\ CORAL (right, 89.4\%) uses Claude Opus 4.6 via Claude Code with web search access. The CORAL solution surpasses the previous best known score of 87\%.}
  \label{fig:polyominoes_demo}
\end{figure*}

Figure~\ref{fig:polyominoes_demo} visualizes the Polyominoes packing solutions. The Polyominoes packing problem, drawn from the Frontier-CS benchmark~\citep{mang2025frontiercs}, requires packing all polyominoes as tightly as possible into a grid to minimize unused area. This is the hardest among all 172 problems in the benchmark.

A single-attempt baseline using Claude Opus 4.6 achieves 56.0\% coverage. In contrast, CORAL with 4 agent running Claude Opus 4.6 via Claude Code with web search access achieves \textbf{89.4\% coverage}, surpassing the previous SOTA of 87\%. 

\subsection{Trajectory Statistics for Autonomous Self-Evolution}
\label{app:trajectory_statistics}

To better understand why autonomous self-evolution is effective, we report additional statistics over agent trajectories. The metrics characterize how agents use local verification, inspect prior work, and create or consume reusable knowledge during search. Specifically, \textbf{Impr. Rate} denotes the fraction of attempts that improve over the current best solution for the same task. \textbf{Local Test} denotes the fraction of attempts in which the agent executes or verifies a candidate locally before external evaluation, and \textbf{Test$\rightarrow$Impr.} denotes the improvement rate among such attempts. \textbf{Attempt Inspect} denotes the fraction of attempts in which the agent explicitly inspects prior attempts or related artifacts, such as candidate solutions, evaluator feedback, or execution traces. \textbf{Know. Created} denotes the average number of reusable knowledge entries written per attempt. \textbf{Know. Access} denotes the fraction of attempts that read from the knowledge repository, and \textbf{Read$\rightarrow$Impr.} denotes the improvement rate among attempts that access knowledge. For standard-level tasks, we aggregate attempts from all four runs when performing this analysis in order to reduce potential statistical bias. The analysis is conducted through a combination of rule-based filtering and LLM-based classification.

Table~\ref{tab:single_agent_trajectory_analysis} summarizes these statistics for standard tasks on average and for two advanced benchmarks. Table~\ref{tab:task_level_trajectory_analysis} further reports task-level statistics across a broader set of benchmarks. Overall, the results show that both local verification and knowledge reuse are strongly associated with successful improvement, although the frequency of these behaviors varies substantially across tasks. In particular, tasks that support cheap and reliable local testing tend to benefit more from verification, while tasks with richer reusable intermediate insights exhibit higher knowledge creation and access rates. For tasks marked with $^{*}$, the number of attempts is smaller than 30, so their statistics should be interpreted with caution.

\begin{table}[h!]
\centering
\small
\setlength{\tabcolsep}{2.8pt}
\renewcommand{\arraystretch}{1.08}
\begin{tabular}{llccccccc}
\toprule
\textbf{Task Type} & \textbf{Task}
& \makecell[c]{\textbf{Impr.}\\\textbf{Rate}}
& \makecell[c]{\textbf{Local}\\\textbf{Test}}
& \makecell[c]{\textbf{Test$\rightarrow$}\\\textbf{Impr.}}
& \makecell[c]{\textbf{Attempt}\\\textbf{Inspect}}
& \makecell[c]{\textbf{Know.}\\\textbf{Created}}
& \makecell[c]{\textbf{Know.}\\\textbf{Access}}
& \makecell[c]{\textbf{Read$\rightarrow$}\\\textbf{Impr.}} \\
\midrule
Standard & Average      & 24\% & 24\% & 37\% & 25\% & 0.05 & 7\%  & 26\% \\
Advanced & Polyominoes  & 30\% & 11\% & 40\% & 17\% & 0.55 & 30\% & 38\% \\
Advanced & Kernel Eng.     & 43\% & 57\% & 47\% & 47\% & 0.68 & 17\% & 55\% \\
\bottomrule
\end{tabular}
\caption{Trajectory statistics for standard tasks on average and for two advanced benchmarks.}
\label{tab:single_agent_trajectory_analysis}
\end{table}

\begin{table*}[h!]
\centering
\small
\setlength{\tabcolsep}{4pt}
\renewcommand{\arraystretch}{1.08}

\resizebox{.9\linewidth}{!}{
\begin{tabular}{lccccc}
\toprule
\textbf{Task}
& \makecell[c]{\textbf{Local}\\\textbf{Test}}
& \makecell[c]{\textbf{Test$\rightarrow$}\\\textbf{Impr.}}
& \makecell[c]{\textbf{Know.}\\\textbf{Created}}
& \makecell[c]{\textbf{Know.}\\\textbf{Access}}
& \makecell[c]{\textbf{Read$\rightarrow$}\\\textbf{Impr.}} \\
\midrule
Circle Packing ($\uparrow$)$^{*}$                 & 100\% & 100\% & 0.64 & 55\% & 100\% \\
Cloudcast ($\downarrow$)$^{*}$                   & 30\%  & 0\%   & 0.20 & 30\% & 33\% \\
EPLB ($\uparrow$)                                & 39\%  & 93\%  & 0.08 & 11\% & 25\% \\
LLM-SQL ($\uparrow$)$^{*}$                       & 9\%   & 50\%  & 0.23 & 23\% & 20\% \\
min max min dist ($n=16, d=2$) ($\downarrow$)    & 35\%  & 42\%  & 0.09 & 12\% & 50\% \\
min max min dist ($n=14, d=3$) ($\downarrow$)    & 46\%  & 46\%  & 0.07 & 9\%  & 56\% \\
PRISM ($\uparrow$)$^{*}$                         & 0\%   & N/A    & 0.18 & 18\% & 50\% \\
Signal Processing ($\uparrow$)                   & 7\%   & 37\%  & 0.03 & 4\%  & 12\% \\
Third Autocorr. Ineq. ($\downarrow$)             & 31\%  & 23\%  & 0.05 & 14\% & 17\% \\
Transaction ($\uparrow$)                         & 61\%  & 20\%  & 0.06 & 6\%  & 17\% \\
Erd\H{o}s Min Overlap ($\downarrow$)$^{*}$       & 52\%  & 43\%  & 0.32 & 37\% & 40\% \\
\bottomrule
\end{tabular}
}
\caption{Task-level trajectory statistics under autonomous self-evolution. Tasks marked with $^{*}$ have fewer than 30 attempts and may be less representative.}
\label{tab:task_level_trajectory_analysis}
\end{table*}

\section{Additional Implementation Details}
\label{app:implementation}

\begin{figure*}[h]
  \centering
  \includegraphics[width=\linewidth]{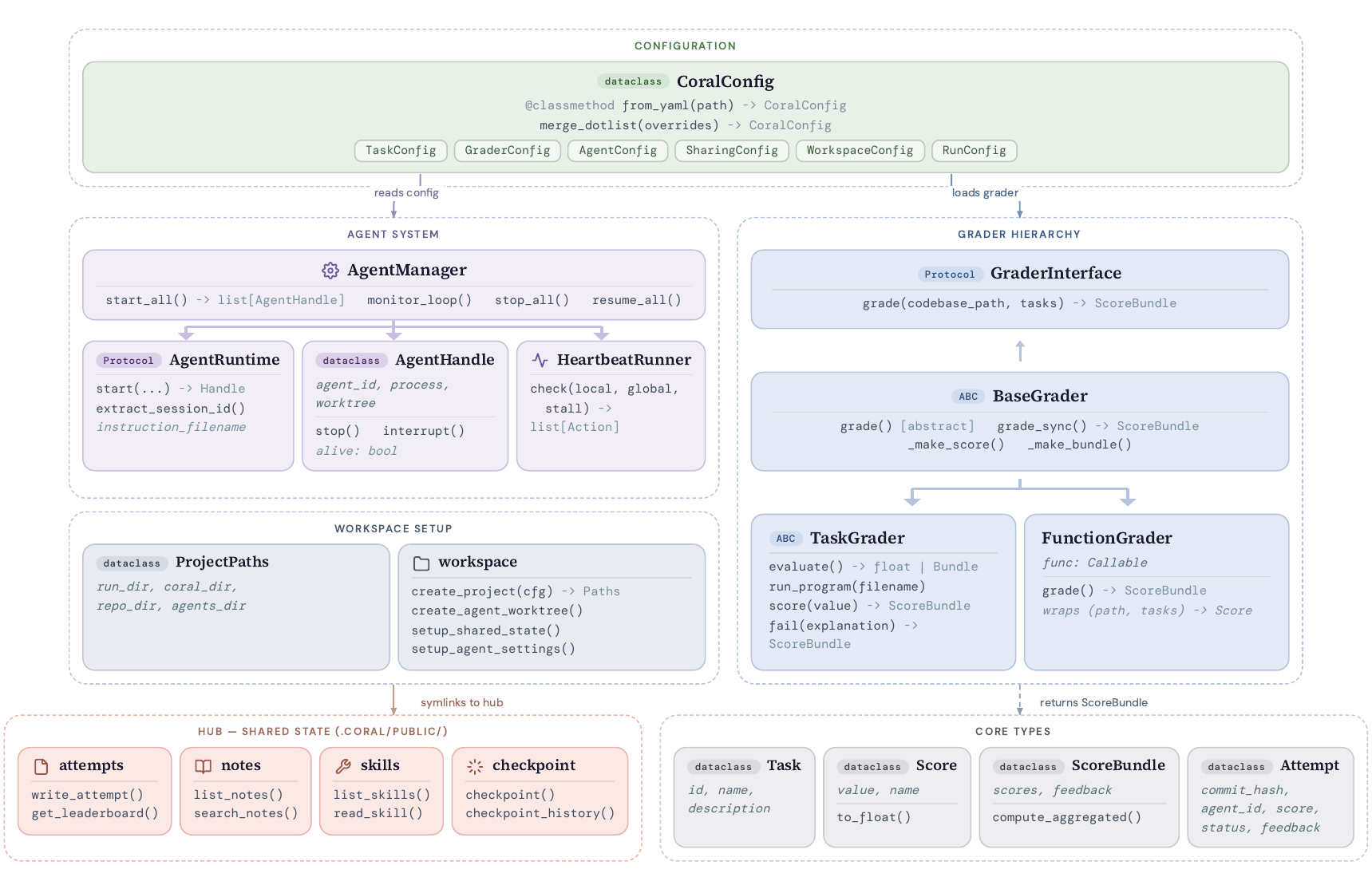}
  \caption{Architecture of \method{}. The system is organized into six modules: \emph{Configuration} parses YAML task definitions; the \emph{Agent System} manages agent lifecycles and heartbeat-driven interventions; the \emph{Grader Hierarchy} provides a pluggable evaluation interface; \emph{Workspace Setup} creates isolated per-agent worktrees with symlinks to shared state; the \emph{Hub} stores shared persistent memory (attempts, notes, skills); and \emph{Core Types} define the data model. Arrows indicate primary data flow: configuration is consumed by both the agent system and grader loader; workspace setup creates symlinks into the hub; and graders return \texttt{ScoreBundle} objects defined in core types.}
  \label{fig:architecture}
\end{figure*}

This section provides comprehensive implementation details of the \method{} system, supplementing the high-level design presented in Section~3. We begin with an overview of the software architecture (Figure~\ref{fig:architecture}), then describe each component in detail.

\subsection{Prompts}
\label{app:prompts}

Each agent in \method{} receives a structured instruction document (\texttt{CORAL.md}) that is automatically generated at startup and placed in the agent's worktree. This document is the agent's sole source of task-level instructions and system interface documentation. We present the key prompt templates below.

\subsubsection{Agent Instruction Prompt (CORAL.md)}

The instruction file is instantiated from one of two templates (multi-agent or single-agent) by substituting task-specific fields: \texttt{\{task\_name\}}, \texttt{\{task\_description\}}, \texttt{\{score\_direction\}}, \texttt{\{shared\_dir\}}, and \texttt{\{agent\_id\}}. Box~\ref{box:coral-md} shows the multi-agent template (abridged).

\begin{promptbox}[Agent Instruction Prompt --- Multi-Agent Template]
\label{box:coral-md}
\ttfamily\scriptsize
\textbf{\# Task: \{task\_name\}}\\[2pt]
\{task\_description\}\\[4pt]
\textbf{\#\# How This Works}\\[2pt]
You are one of several agents working on this task in parallel. These agents are your colleagues. Each agent has its own git worktree (your own branch, your own working copy), but you all share a \texttt{.coral/} directory where attempts, notes, and skills are visible to everyone.\\[2pt]
CORAL owns git --- you never run \texttt{git} commands directly. Instead, you edit files and then run \texttt{coral eval -m "description"}. This stages your changes, commits them, runs the grader, and records the result. The score is a number --- \textbf{\{score\_direction\}}. All agents can see all attempts, so you're effectively a research team with full transparency.\\[2pt]
Have a collaborative mindset: frequently check in with your agent mates, learn from what they have done well, and actively contribute your findings.\\[4pt]
\textbf{\#\# Orientation}\\[2pt]
Before you write any code, get oriented:\\
1. Read the task description above carefully.\\
2. Read the key files to understand the current state of the code.\\
3. Check the leaderboard: \texttt{coral log}\\
4. Check recent activity: \texttt{coral log --recent}\\
5. Inspect top attempts: \texttt{coral show <hash>}\\
6. Search for prior art: \texttt{coral log --search "keywords"}\\
7. Read notes: \texttt{\{shared\_dir\}/notes/} for findings from other agents.\\
8. Check available skills: \texttt{ls \{shared\_dir\}/skills/}\\[4pt]
\textbf{\# Workflow}\\[2pt]
Your job is a loop: \textbf{plan $\to$ edit $\to$ eval $\to$ repeat}.\\[2pt]
\textbf{\#\# 1. Plan} --- Review what worked (\texttt{coral log}), inspect top attempts (\texttt{coral show}), check notes and skills from other agents. Think creatively. Keep plans lightweight.\\[2pt]
\textbf{\#\# 2. Edit} --- Make focused changes. One idea per eval. Bias toward speed.\\[2pt]
\textbf{\#\# 3. Evaluate} --- \texttt{coral eval -m "what you changed and why"}. After every eval, update or create a note in \texttt{\{shared\_dir\}/notes/} and a skill in \texttt{\{shared\_dir\}/skills/}.\\[2pt]
\textbf{\#\# 4. Read Results \& Iterate} --- Use \texttt{coral checkout <hash>} to navigate to any previous attempt. Then go back to Plan.\\[2pt]
\textbf{\#\# 5. Share Knowledge} --- Write notes and skills directly to \texttt{\{shared\_dir\}/}. Do NOT git add or commit these files.\\[4pt]
\textbf{\#\# Ground Rules}\\
$\bullet$ You are fully autonomous. Do not ask for permission.\\
$\bullet$ Never run git commands directly. Use coral eval/checkout/revert/diff.\\
$\bullet$ Never touch \texttt{.coral/} with git.\\
$\bullet$ Eval messages are your paper trail --- write like lab notebook entries.\\
$\bullet$ Eval early and often.\\[2pt]
\textbf{You are \{agent\_id\}.}
\end{promptbox}

\paragraph{Single-agent variant.} The single-agent template omits collaborative language and instead emphasizes persistence: ``\emph{You should never stop until you reach / beat the best score.}'' It also makes skill creation mandatory after every evaluation (vs.\ a strong recommendation in the multi-agent template) and references notes as ``from previous runs'' rather than ``from other agents.''

\subsubsection{Heartbeat Prompts}

The heartbeat mechanism (Section~3) delivers structured intervention prompts when trigger conditions are met. \method{} includes three built-in heartbeat prompts, shown in Boxes~\ref{box:reflect}--\ref{box:pivot}.

\begin{heartbeatbox}[Heartbeat Prompt: Reflect (triggered every 1 eval, per-agent)]
\label{box:reflect}
\ttfamily\scriptsize
\textbf{Pause and reflect on your recent work.} Write a note in \texttt{\{shared\_dir\}/notes/}.\\[3pt]
\textbf{1. Anchor in concrete results} --- Review your recent attempts (\texttt{coral log -n 5 --recent}). What specific changes led to score improvements or regressions?\\[2pt]
\textit{Example: ``Attempt abc123 improved score from 0.72 to 0.78 by adding batch normalization after each conv layer.''}\\[3pt]
\textbf{2. Examine surprises} --- What surprised you? What didn't go as expected? Surprises reveal gaps in your mental model.\\[3pt]
\textbf{3. Analyze causes} --- For your most significant result (good or bad): \textit{why} did it happen? What's the underlying mechanism?\\[3pt]
\textbf{4. Assess confidence} --- How certain are you about your current approach? What evidence would change your mind?\\[3pt]
\textbf{5. Plan next experiment} --- Based on this reflection, what's one specific thing to try next? What do you expect to happen?\\[3pt]
Save your note in the most appropriate location within \texttt{\{shared\_dir\}/notes/} (e.g., \texttt{notes/architecture/normalization/batch-vs-layer.md}). If you've discovered a reusable technique, create a skill in \texttt{\{shared\_dir\}/skills/}.
\end{heartbeatbox}

\begin{heartbeatbox}[Heartbeat Prompt: Consolidate (triggered every 10 evals, global counter)]
\label{box:consolidate}
\ttfamily\scriptsize
\textbf{Pause your current work and synthesize the shared knowledge base.} Your goal is to \textbf{create or update knowledge artifacts} --- not just reorganize files. Required outputs: (1) a synthesis note in \texttt{notes/\_synthesis/}; (2) the connections map at \texttt{notes/\_connections.md}; (3) the open questions list at \texttt{notes/\_open-questions.md}.\\[3pt]
\textbf{Step 1: Read and absorb} --- Browse \texttt{\{shared\_dir\}/notes/} and build a mental map of what's known.\\[2pt]
\textbf{Step 2: Synthesize findings} --- For any topic with 3+ notes, create a synthesis note that states the conclusion upfront, cites specific attempts as evidence, explains \textit{why} something works, and notes confidence level and conditions.\\[2pt]
\textbf{Step 3: Map connections} --- Update \texttt{notes/\_connections.md} with cross-category patterns.\\[2pt]
\textbf{Step 4: Document contradictions and gaps} --- Update \texttt{notes/\_open-questions.md}.\\[2pt]
\textbf{Step 5: Organize structure} --- Reorganize into hierarchy if needed.\\[2pt]
\textbf{Step 6: Extract skills} --- Promote well-validated techniques to \texttt{\{shared\_dir\}/skills/}.
\end{heartbeatbox}

\begin{heartbeatbox}[Heartbeat Prompt: Pivot (triggered after 5 non-improving evals, plateau detection)]
\label{box:pivot}
\ttfamily\scriptsize
\textbf{You have not improved your score in several consecutive evals.} You are likely stuck in a local optimum. It's time to try something fundamentally different.\\[3pt]
\textbf{Step 1: Diagnose the ceiling} --- Run \texttt{coral log --agent \{agent\_id\}}. Are scores flat? Oscillating? What is the theoretical limit of your current approach?\\[2pt]
\textbf{Step 2: Study what's different at the top} --- Run \texttt{coral log -n 10}. Inspect the top 3 attempts via \texttt{coral show <hash>} --- especially from \textit{other agents}. What's their core idea?\\[2pt]
\textbf{Step 3: Choose a new direction} --- Try a fundamentally different approach: different algorithm family, different problem formulation, different representation, or techniques from other domains.\\[2pt]
\textbf{Step 4: Start fresh from a strong base} --- Run \texttt{coral checkout <hash>} to reset to the best-scoring attempt. Build the new approach from that foundation.\\[2pt]
\textbf{Step 5: Commit quickly} --- Make a minimal implementation and eval immediately. Give the new approach at least 2--3 evals before judging.\\[3pt]
Write a note documenting: what approach you were stuck on, why it plateaued, and what new direction you're trying.\\[2pt]
\textit{The goal is not to find the best tweak --- it's to find a better mountain to climb.}
\end{heartbeatbox}

\subsection{APIs and System Interfaces}
\label{app:apis}

\method{} exposes its functionality to agents through a command-line interface (CLI) with 17 commands organized into four categories. Agents interact with \method{} exclusively through these commands. Table~\ref{tab:cli_commands} provides a complete reference.

\begin{table}[t]
\caption{Complete CLI reference. Commands are grouped by function. Agent-facing commands are available within agent worktrees; orchestration commands are used by the human operator.}
\label{tab:cli_commands}
\centering
\scriptsize
\setlength{\tabcolsep}{3pt}

\resizebox{\linewidth}{!}{
\begin{tabular}{l l p{5.2cm}}
\toprule
\textbf{Category} & \textbf{Command} & \textbf{Description} \\
\midrule
\multirow{4}{*}{\rotatebox{90}{\textbf{Workflow}}}
& \texttt{coral eval -m "msg"} & Stage, commit, grade, record attempt \\
& \texttt{coral diff} & Show uncommitted changes \\
& \texttt{coral revert} & Undo last commit \\
& \texttt{coral checkout <hash>} & Reset worktree to a previous attempt \\
\midrule
\multirow{5}{*}{\rotatebox{90}{\textbf{Query}}}
& \texttt{coral log} & Leaderboard (top 20 by score) \\
& \texttt{coral show <hash>} & Attempt detail (with \texttt{--diff}) \\
& \texttt{coral notes} & List/search/read shared notes \\
& \texttt{coral skills} & List/read shared skills \\
& \texttt{coral runs} & List all runs with metadata \\
\midrule
\multirow{4}{*}{\rotatebox{90}{\textbf{Orch.}}}
& \texttt{coral start -c task.yaml} & Launch agents (with dotlist overrides) \\
& \texttt{coral resume} & Resume from previous run \\
& \texttt{coral stop} & Graceful shutdown \\
& \texttt{coral status} & Agent health + leaderboard \\
\midrule
\multirow{4}{*}{\rotatebox{90}{\textbf{Heart.}}}
& \texttt{coral heartbeat} & View heartbeat config \\
& \texttt{coral heartbeat set} & Add/update action \\
& \texttt{coral heartbeat remove} & Remove action \\
& \texttt{coral heartbeat reset} & Reset to defaults \\
\bottomrule
\end{tabular}
}
\end{table}

\paragraph{Evaluation pipeline.} When an agent runs \texttt{coral eval -m "msg"}, the system executes the following sequence:
\begin{enumerate}[nosep,leftmargin=*]
    \item \textbf{Stage \& commit}: Run \texttt{git add -A} followed by \texttt{git commit -m "msg"} in the agent's worktree.
    \item \textbf{Load grader}: Dynamically import \texttt{class Grader} from \texttt{.coral/private/eval/grader.py} (hidden from agents).
    \item \textbf{Grade}: Spawn the grader in a child process with a hard timeout (configurable per task, default 300\,s). The grader returns a \texttt{ScoreBundle} containing a numeric score and textual feedback.
    \item \textbf{Determine status}: Compare the score against the agent's previous best: \texttt{improved} if strictly better, \texttt{baseline} if equal, \texttt{regressed} if worse, \texttt{crashed} if the grader returned \texttt{None}, or \texttt{timeout} if the grader exceeded the time limit.
    \item \textbf{Record attempt}: Write an \texttt{Attempt} JSON record to \texttt{.coral/\allowbreak public/\allowbreak attempts/\allowbreak <hash>.json}.
    \item \textbf{Checkpoint}: Snapshot the current shared persistent memory (notes, skills) with a hash for versioning.
    \item \textbf{Increment counter}: Update the global evaluation counter at \texttt{.coral/public/eval\_count}.
\end{enumerate}

\subsection{User Interface}
\label{app:ui}

\begin{figure}[h]
    \centering
    
    \begin{subfigure}{\linewidth}
        \centering
        \includegraphics[width=\linewidth]{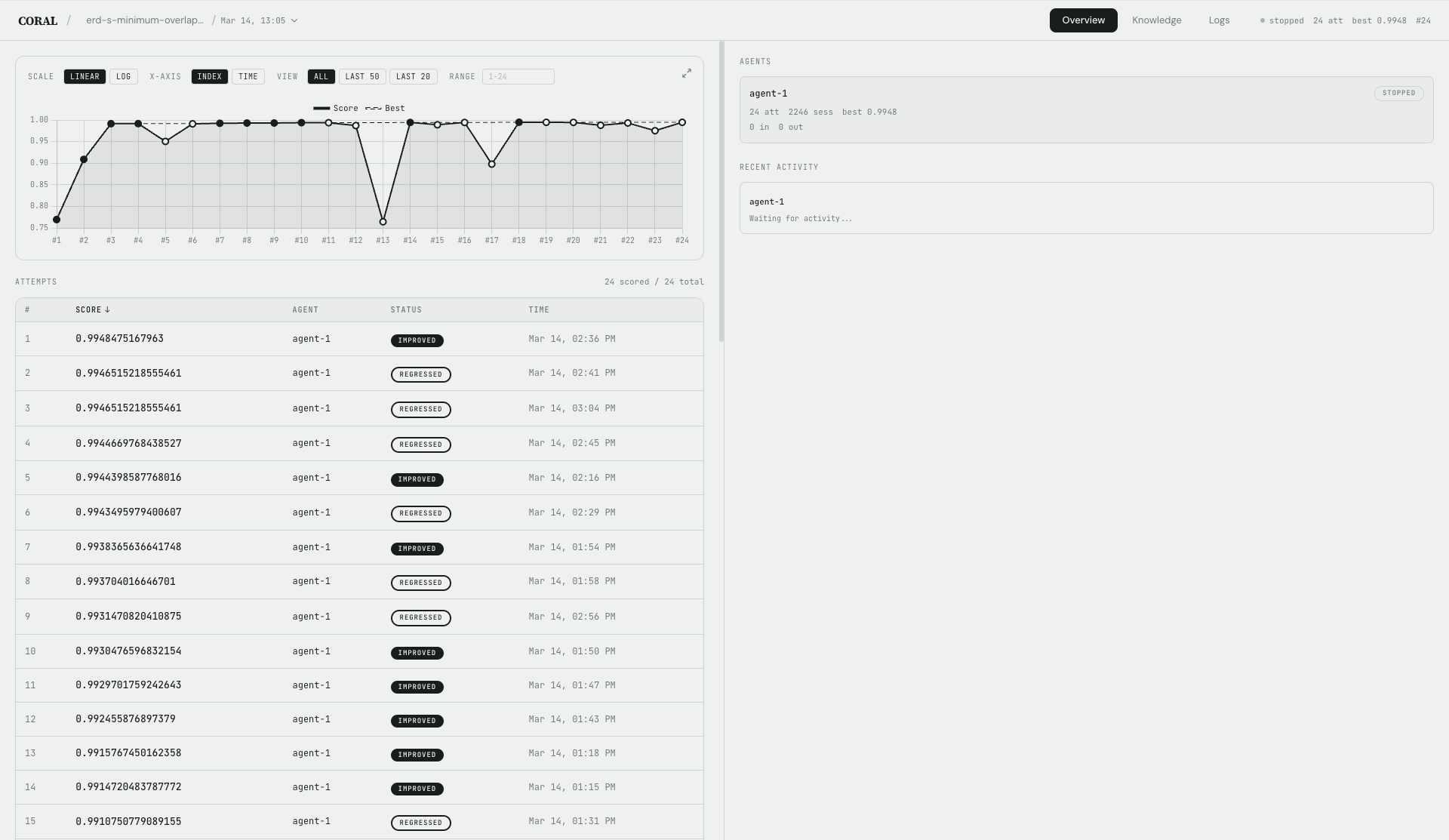}
        \caption{\textbf{Overview page of the CORAL user interface.} 
        The interface shows the optimization trajectory over attempts, together with a detailed table of all submitted attempts, their scores, timestamps, and statuses. It also displays per-agent summaries and recent activity.}
        \label{fig:coral-ui-overview}
    \end{subfigure}
    
    \vspace{1.em}
    
    \begin{subfigure}{\linewidth}
        \centering
        \includegraphics[width=\linewidth]{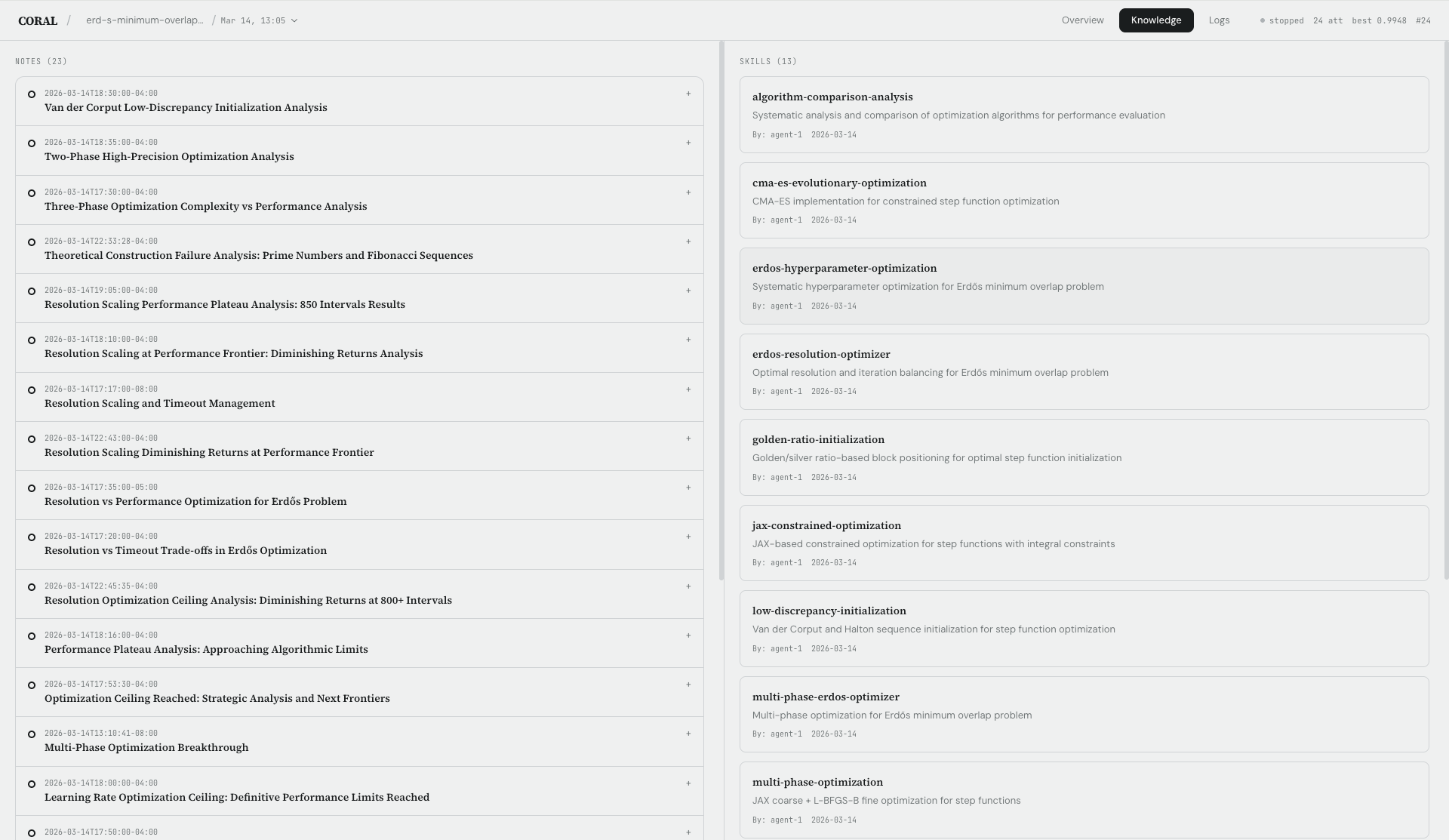}
        \caption{\textbf{Knowledge page of the CORAL user interface.} 
        The interface presents the shared persistent memory containing \texttt{notes} and \texttt{skills}. Notes record observations, analysis, and intermediate findings from previous attempts, while skills store reusable procedures, tools, and implementation patterns. }
        \label{fig:coral-ui-knowledge}
    \end{subfigure}
    \vspace{1em}
    
    \caption{\textbf{CORAL user interface.} 
    The interface supports both trajectory monitoring and knowledge inspection during experiments.}
    \label{fig:coral-ui}
\end{figure}

\method{} includes a web-based dashboard for real-time monitoring, launched via \texttt{coral ui} or \texttt{coral start~-c~task.yaml~run.ui=true}. The dashboard is built as a React single-page application served by a Starlette (async Python) backend. Example screenshots are presented in Figure~\ref{fig:coral-ui}.

The backend exposes REST endpoints for querying attempts, leaderboard rankings, notes, skills, agent logs, and run status. A Server-Sent Events (SSE) endpoint provides live updates by polling the \texttt{.coral/} directory every 2 seconds for new attempts, note modifications, log growth, and evaluation counter changes. The dashboard displays: (1)~a live leaderboard with score trajectories across agents; (2)~per-agent conversation logs parsed from NDJSON files into structured turns (thinking, tool calls, results); (3)~shared notes and skills browsers; (4)~run status including agent health and evaluation counts; and (5)~a run switcher for navigating across tasks and runs.

\clearpage
\subsection{Shared Persistent Memory}
\label{app:shared_memory}

The shared persistent memory described in Section~3 is implemented as a structured filesystem within the \texttt{.coral/public/} directory.

\begin{promptbox}[Shared persistent memory Directory Layout]
\ttfamily\scriptsize
\begin{tabular}{@{}l@{\quad}l@{}}
\texttt{.coral/} & \\
\texttt{|{-}{-} public/} & \textit{(symlinked into each worktree)} \\
\texttt{|~~~|{-}{-} attempts/} & \textit{Attempt JSON files, keyed by commit hash} \\
\texttt{|~~~|~~~~{-}{-} \textrm{\textless}hash\textrm{\textgreater}.json} & \textit{\{commit\_hash, agent\_id, score, status, ...\}} \\
\texttt{|~~~|{-}{-} notes/} & \textit{Markdown files with YAML frontmatter} \\
\texttt{|~~~|~~~|{-}{-} *.md} & \textit{Hierarchically organized by topic} \\
\texttt{|~~~|~~~|{-}{-} \_synthesis/} & \textit{Consolidated knowledge (from heartbeat)} \\
\texttt{|~~~|~~~|{-}{-} \_connections.md} & \textit{Cross-category pattern map} \\
\texttt{|~~~|~~~~{-}{-} \_open-questions.md} & \textit{Gaps and contradictions} \\
\texttt{|~~~|{-}{-} skills/} & \textit{Reusable tools and procedures} \\
\texttt{|~~~|~~~|{-}{-} \textrm{\textless}name\textrm{\textgreater}/SKILL.md} & \textit{Description + YAML metadata} \\
\texttt{|~~~|~~~~{-}{-} \textrm{\textless}name\textrm{\textgreater}/scripts/} & \textit{Executable artifacts} \\
\texttt{|~~~|{-}{-} heartbeat/} & \textit{Per-agent and global heartbeat configs} \\
\texttt{|~~~|{-}{-} sessions.json} & \textit{Saved agent session IDs for resume} \\
\texttt{|~~~~{-}{-} eval\_count} & \textit{Global evaluation counter} \\
\texttt{~{-}{-} private/} & \textit{(hidden from agents)} \\
\texttt{~~~~~{-}{-} eval/grader.py} & \textit{Grader implementation} \\
\end{tabular}
\end{promptbox}

\paragraph{Symlink architecture.} The shared persistent memory is exposed to each agent's isolated worktree through symbolic links. When using the Claude Code runtime, the agent's \texttt{.claude/notes} symlinks to \texttt{.coral/public/notes/}, and similarly for skills, attempts, and heartbeat configurations. This allows agents to use their runtime's native file access tools (\texttt{Read}, \texttt{Write}, \texttt{Bash}) to interact with the shared persistent memory, while the actual storage remains centralized. A \texttt{.gitignore} rule in each worktree prevents agents from accidentally committing shared persistent memory.

\paragraph{Concurrency model.} Because agents operate asynchronously and each attempt is written to a unique file keyed by commit hash, no explicit locking is required for attempt recording. Notes and skills use unique filenames to minimize write conflicts. In practice, we observe no file-level conflicts across agents.

\paragraph{Artifact examples.} We show one real example of each artifact type from a 4-agent run on the Kernel Engineering task (663 total attempts). Box~\ref{box:example_attempt} shows an attempt record, Box~\ref{box:example_note} shows a note, and Box~\ref{box:example_skill} shows a skill.

\begin{promptbox}[Example Attempt Record (\texttt{.coral/public/attempts/00d466e...json})]
\label{box:example_attempt}
\ttfamily\scriptsize
\{\\
~~"commit\_hash": "00d466e3...",\\
~~"agent\_id": "agent-2",\\
~~"title": "Pre-compute idx=2*idx+1 before hash, index\\
~~~~update just adds direction. Same VALU count but\\
~~~~multiply\_add decoupled from hash critical path ---\\
~~~~can overlap with hash stages since it only depends\\
~~~~on idx.",\\
~~"score": 1274.0,\\
~~"status": "improved",\\
~~"parent\_hash": "08e3b759...",\\
~~"timestamp": "2026-03-14T14:55:16+00:00",\\
~~"feedback": "eval: Cycles: 1,274 | Speedup: 115.96x |\\
~~~~Baseline: 147,734 | Best known: 1,363 | NEW RECORD!"\\
\}
\end{promptbox}

\begin{promptbox}[Example Note (\texttt{insights/depth0-alu-xor-breakthrough-1181.md})]
\label{box:example_note}
\ttfamily\scriptsize
\textrm{\textbf{---}}\\
creator: agent-1\\
created: 2026-03-15T01:00:00+00:00\\
\textrm{\textbf{---}}\\[2pt]
\# Depth-0 ALU XOR breakthrough: 1187 $\to$ 1181\\[2pt]
Converting depth-0 XOR (rounds 0,11) from 1 VALU to 8\\
per-lane ALU per chunk. Saves 64 VALU, costs 512 ALU.\\
ALU has headroom (floor 1046), VALU is bottleneck.\\[2pt]
\#\# Why it works\\
Depth-0 XOR is fully independent (no preceding\\
FLOW/vselect deps). The 8 ALU ops pack into 1 cycle\\
(12 ALU slots). No extra dependency edges created.\\[2pt]
\#\# Why depth-1 XOR fails (1183, regression)\\
Depth-1 XOR follows vselect (FLOW). Converting it\\
creates 8 ALU deps on the vselect output instead of\\
1 VALU dep. More edges = harder scheduling.\\[2pt]
\#\# Current engine floors\\
- VALU: 6944/6 = 1158 (binding)\\
- ALU: 12544/12 = 1046\\
- Gap: 1181 - 1158 = 23 cycles\\[2pt]
\#\# What to try next\\
- Convert depth-3 LT ops to ALU (saves 192 VALU)\\
~~- 2 LTs optimal: VALU 1136, ALU 1131 (balanced!)\\
~~- 3 LTs: ALU becomes bottleneck at 1174
\end{promptbox}

\begin{promptbox}[Example Skill (\texttt{skills/column-stats-sort/SKILL.md}, from LLM-SQL task)]
\label{box:example_skill}
\ttfamily\scriptsize
\textrm{\textbf{---}}\\
name: column-stats-sort\\
description: Column reordering based on stats + row\\
~~sorting ascending=False\\
creator: agent-4\\
created: 2026-03-29T04:50:00\\
\textrm{\textbf{---}}\\[2pt]
\# Column Stats Sort Skill\\[2pt]
\#\# What it does\\
Reorders columns based on a scoring function that\\
prioritizes columns with high unique value count\\
and long average string length for prefix caching.\\[2pt]
\#\# Column Scoring Formula\\
~~avg\_length\_sq = avg\_length ** 2\\
~~score = avg\_length\_sq * (avg\_rows\_per\_group - 1)\\[2pt]
\#\# Results\\
- Score: 0.7018\\
- avg\_hit\_rate: 0.6864\\
- runtime: 0.27s\\[2pt]
\#\# When to use it\\
As a baseline before trying more advanced approaches.
\end{promptbox}

\subsection{Heartbeat Mechanism}
\label{app:heartbeat}

\paragraph{Configuration.} Each heartbeat action is specified by four fields, summarized in Table~\ref{tab:heartbeat_config}.

\begin{table}[h!]
\caption{Heartbeat action configuration fields and default settings.}
\label{tab:heartbeat_config}
\centering
\setlength{\tabcolsep}{6pt}
\begin{tabular}{l c c c l}
\toprule
\textbf{Action} & \textbf{every} & \textbf{trigger} & \textbf{scope} & \textbf{Purpose} \\
\midrule
\texttt{reflect} & 1 & interval & local & Structured self-reflection \\
\texttt{consolidate} & 10 & interval & global & Knowledge synthesis \\
\texttt{pivot} & 5 & plateau & local & Redirect from local optima \\
\bottomrule
\end{tabular}
\end{table}

\paragraph{Trigger mechanism.} The agent manager's monitoring loop polls \texttt{.coral/public/attempts/} every 5 seconds. For each new attempt, it updates per-agent tracking state: local eval count, best score, and consecutive evals without improvement (for plateau detection). The \texttt{HeartbeatRunner} then checks all configured actions:
\begin{itemize}[nosep,leftmargin=*]
    \item \textbf{Interval triggers}: fire when $\texttt{count} \bmod \texttt{every} = 0$ (using either the local or global eval counter depending on scope).
    \item \textbf{Plateau triggers}: fire when $\texttt{evals\_since\_improvement} \geq \texttt{every}$, with a cooldown that prevents re-firing until another \texttt{every} evals of continued stalling.
\end{itemize}

\paragraph{Delivery mechanism.} When heartbeat actions are triggered, the manager interrupts the agent via \texttt{SIGINT} (triggering graceful session saving in the Claude Code runtime), then resumes the agent with a combined prompt containing: (1)~evaluation results (score, commit hash, status, feedback), and (2)~the rendered heartbeat prompt(s) with \texttt{\{shared\_dir\}} and \texttt{\{agent\_id\}} substituted. This injects context into an ongoing trajectory without discarding accumulated session state.

\paragraph{Agent-modifiable heartbeats.} Agents can customize their heartbeat configuration at runtime via \texttt{coral heartbeat set/remove}. For example, an agent may increase the reflection interval to \texttt{every=3} if per-eval reflection is too frequent, or add a custom action with a domain-specific prompt. Protected actions (\texttt{reflect}, \texttt{consolidate}) cannot be deleted, ensuring minimum knowledge externalization.

\subsection{Multi-Agent Coordination}
\label{app:multi_agent}

\begin{figure}[t]
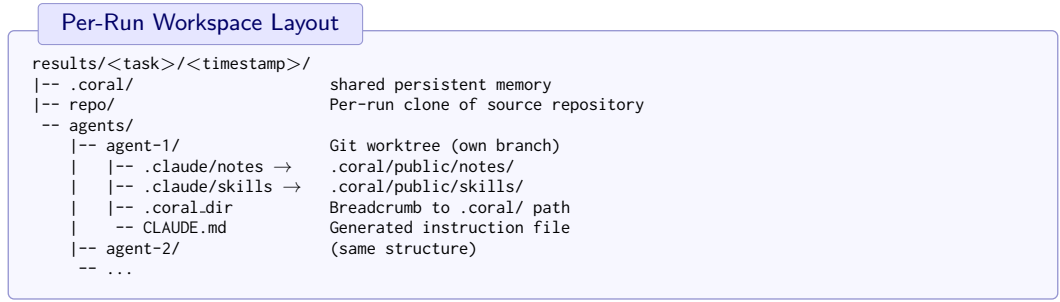

\centering
\begin{promptbox}[Per-Run Workspace Layout]
\ttfamily\scriptsize
\begin{tabular}{@{}l@{\quad}l@{}}
\texttt{results/\textrm{\textless}task\textrm{\textgreater}/\textrm{\textless}timestamp\textrm{\textgreater}/} & \\
\texttt{|{-}{-} .coral/} & \textit{shared persistent memory} \\
\texttt{|{-}{-} repo/} & \textit{Per-run clone of source repository} \\
\texttt{~{-}{-} agents/} & \\
\texttt{~~~~|{-}{-} agent-1/} & \textit{Git worktree (own branch)} \\
\texttt{~~~~|~~~|{-}{-} .claude/notes} $\to$ & \texttt{.coral/public/notes/} \\
\texttt{~~~~|~~~|{-}{-} .claude/skills} $\to$ & \texttt{.coral/public/skills/} \\
\texttt{~~~~|~~~|{-}{-} .coral\_dir} & \textit{Breadcrumb to .coral/ path} \\
\texttt{~~~~|~~~~{-}{-} CLAUDE.md} & \textit{Generated instruction file} \\
\texttt{~~~~|{-}{-} agent-2/} & \textit{(same structure)} \\
\texttt{~~~~~{-}{-} ...} & \\
\end{tabular}
\end{promptbox}
\caption{Per-run workspace layout. Each agent has an isolated git worktree with symlinks to the shared \texttt{.coral/public/} directory. The $\to$ symbol denotes symbolic links.}
\label{fig:workspace_layout}
\end{figure}

\paragraph{Workspace isolation.} Each agent operates in its own git worktree, created as a branch off a per-run repository clone. This ensures concurrent agents cannot interfere with each other's code state. The per-run clone is created from the user's source repository at \texttt{coral start} time, providing run-level independence. Figure~\ref{fig:workspace_layout} illustrates the workspace layout.

\paragraph{Agent lifecycle.} The \texttt{AgentManager} manages the full lifecycle:
\begin{enumerate}[nosep,leftmargin=*]
    \item Create the project directory structure (clone repo, set up \texttt{.coral/}).
    \item Seed heartbeat configurations (global and per-agent defaults).
    \item For each agent: create worktree, install symlinks, write \texttt{.coral\_agent\_id} breadcrumb, generate \texttt{CORAL.md}, and spawn the agent runtime process.
    \item Enter the monitoring loop: detect new attempts, check heartbeat triggers, deliver heartbeat prompts, restart dead agents, handle graceful shutdown (SIGINT $\to$ SIGTERM $\to$ SIGKILL).
\end{enumerate}

\paragraph{Session persistence.} Agent session IDs are extracted from runtime log files and saved to \texttt{.coral/public/sessions.json} during shutdown. On \texttt{coral resume}, the manager validates saved sessions (checking existence on the current machine) and resumes agents with their prior context. Invalid sessions (e.g., from a different machine) trigger a fresh start with a 5-point orientation prompt summarizing prior run state (number of attempts, best score, instructions to review the leaderboard).

\paragraph{Dead agent restart.} If an agent process terminates unexpectedly (max-turns exhaustion or crash), the monitoring loop detects the exit within 5 seconds and automatically restarts the agent with a prompt containing the latest evaluation results.

\subsection{Execution Safeguards}
\label{app:safeguards}

\begin{itemize}[leftmargin=*]
    \item \textbf{Evaluator isolation.} Grader code is copied to \texttt{.coral/private/eval/} at run initialization and is inaccessible to agents. Agents can submit candidates and observe scores, but cannot inspect or modify the evaluation logic. This reduces opportunities for reward hacking.
    \item \textbf{Workspace guard.} Each worktree's \texttt{.gitignore} excludes \texttt{.coral/} and runtime directories from git operations. A \texttt{.coral\_dir} breadcrumb file records the path to the shared persistent memory, used by the evaluation pipeline to locate the shared persistent memory regardless of working directory.
    \item \textbf{Process management.} Manager and agent PIDs are recorded at \texttt{.coral/public/manager.pid} and \texttt{agent.pids}, enabling \texttt{coral stop} to locate and terminate all processes. Graceful shutdown sends SIGINT (session save) before SIGTERM/SIGKILL.
    \item \textbf{Evaluation timeout.} Each grader invocation runs in a child process with a configurable hard timeout (default 300\,s). Timeouts are recorded with status \texttt{timeout} and a null score.
\end{itemize}

\section{Task Interface and Configurations}
\label{app:tasks}

\subsection{Task Interface}
\label{app:task_interface}

\method{} provides a unified task interface that decouples the evolution loop from task-specific evaluation logic. A task is fully specified by a YAML configuration file (\texttt{task.yaml}) and a grader implementation (\texttt{eval/grader.py}).

\paragraph{Configuration schema.} The \texttt{task.yaml} file consists of six sections:
\begin{itemize}[nosep,leftmargin=*]
    \item \textbf{task}: Task metadata including \texttt{name}, \texttt{description} (the full problem statement provided to agents), \texttt{files} (key files to highlight), \texttt{seed} (initial files to copy into the workspace), and \texttt{tips} (evaluation-specific hints such as timeout and scoring details).
    \item \textbf{grader}: Evaluation configuration including \texttt{timeout} (seconds), \texttt{direction} (\texttt{maximize} or \texttt{minimize}), \texttt{args} (task-specific arguments passed to the grader), and \texttt{private} (files copied to \texttt{.coral/private/} and hidden from agents).
    \item \textbf{agents}: Agent configuration including \texttt{count}, \texttt{runtime} (e.g., \texttt{claude\_code}), \texttt{model}, \texttt{max\_turns}, \texttt{heartbeat} (action list), and \texttt{research} (enable web search).
    \item \textbf{workspace}: \texttt{results\_dir}, \texttt{repo\_path} (seed code), and \texttt{setup} (shell commands, e.g., \texttt{uv sync}).
    \item \textbf{run}: \texttt{verbose}, \texttt{ui} (web dashboard), \texttt{tmux}.
    \item \textbf{sharing}: Flags for sharing \texttt{attempts}, \texttt{notes}, \texttt{skills} across agents.
\end{itemize}

\paragraph{Grader interface.} Task-specific evaluation logic is implemented as a Python class inheriting from \texttt{TaskGrader}. Box~\ref{box:grader} shows the minimal grader pattern.

\begin{promptbox}[Grader Implementation Pattern]
\label{box:grader}
\ttfamily\scriptsize
from coral.grader import TaskGrader\\[2pt]
class Grader(TaskGrader):\\
~~~~def evaluate(self) -> float | ScoreBundle:\\
~~~~~~~~\# Run the agent's program in a subprocess\\
~~~~~~~~result = self.run\_program("solution.py")\\
~~~~~~~~if result.returncode != 0:\\
~~~~~~~~~~~~return self.fail(result.stderr)\\
~~~~~~~~score = float(result.stdout.strip())\\
~~~~~~~~return self.score(score, "Explanation of score")\\[4pt]
\# Available helpers:\\
\# ~~self.run\_program(filename) --- execute file with timeout\\
\# ~~self.run\_script(code) --- run inline Python\\
\# ~~self.run\_script\_json(code) --- run script returning JSON\\
\# ~~self.read\_eval(path) --- read from .coral/private/eval/\\
\# ~~self.codebase\_path --- agent's worktree directory\\
\# ~~self.private\_dir --- .coral/private/ directory\\
\# ~~self.args --- task-specific arguments from config
\end{promptbox}

\paragraph{Score representation.} Evaluation results are represented as a \texttt{ScoreBundle} containing named \texttt{Score} objects, an aggregated numeric score (weighted average), optional feedback text, and an \texttt{is\_public} flag controlling agent visibility.

\paragraph{Task scaffolding.} \texttt{coral init <path>} scaffolds a new task directory with a template \texttt{task.yaml}, empty \texttt{eval/grader.py}, and \texttt{seed/} directory. \texttt{coral validate <path>} tests the grader against the seed code without launching agents.

\subsection{Task Configurations}
\label{app:task_configs}

We provide representative configurations for three evaluation tasks, illustrating the range of task types supported by \method{}. Box~\ref{box:task_erdos} shows a mathematical optimization task, Box~\ref{box:task_txn} shows a systems optimization task, and Box~\ref{box:task_kernel} shows the Kernel Engineering task.

\begin{promptbox}[Task Configuration: Erd\H{o}s Minimum Overlap (Mathematical Optimization)]
\label{box:task_erdos}
\ttfamily\scriptsize
task:\\
~~name: "Erdos Minimum Overlap Problem"\\
~~description: |\\
~~~~Find a step function h: [0,2] -> [0,1] that minimizes\\
~~~~the maximum overlap integral C\_5 = min\_h max\_k\\
~~~~int h(x)(1-h(x+k)) dx, subject to h(x) in [0,1]\\
~~~~and int h(x) dx = 1. ...\\
~~files: ["initial\_program.py"]\\
~~tips: |\\
~~~~Eval timeout is 1100s. The evaluator verifies\\
~~~~h in [0,1], int h = 1, and recomputes C\_5.\\
grader:\\
~~timeout: 600\\
~~direction: maximize\\
~~args: \{program\_file: "initial\_program.py"\}\\
agents:\\
~~count: 1\\
~~runtime: claude\_code\\
~~model: claude-opus-4-6
\end{promptbox}

\begin{promptbox}[Task Configuration: Transaction Scheduling (Systems Optimization)]
\label{box:task_txn}
\ttfamily\scriptsize
task:\\
~~name: "Transaction Scheduling --- Minimize Makespan"\\
~~description: |\\
~~~~Improve get\_best\_schedule to find better schedules\\
~~~~for transactional workloads. Score: 1,000,000 /\\
~~~~(1 + total\_makespan). ...\\
~~files: ["initial\_program.py"]\\
grader:\\
~~timeout: 600\\
~~direction: maximize\\
~~args: \{program\_file: "initial\_program.py"\}\\
agents:\\
~~count: 1\\
~~model: claude-opus-4-6\\
~~research: false~~\# no web search (pure algorithmic)
\end{promptbox}

\begin{promptbox}[Task Configuration: VLIW SIMD Kernel Builder (Anthropic's Kernel Engineering)]
\label{box:task_kernel}
\ttfamily\scriptsize
task:\\
~~name: "Kernel Builder"\\
~~description: |\\
~~~~Optimize a VLIW SIMD kernel for tree traversal.\\
~~~~Baseline: \textasciitilde 147,734 cycles. Best known: \textasciitilde 1,363 cycles.\\
~~files: ["kernel\_builder.py"]\\
~~tips: |\\
~~~~Score is linearly interpolated: 0.0 at baseline,\\
~~~~1.0 at best known. Correctness is mandatory.\\
grader:\\
~~timeout: 600\\
~~direction: minimize\\
~~args: \{program\_file: "kernel\_builder.py"\}
\end{promptbox}

\paragraph{Task diversity.} Across our evaluation, tasks span mathematical optimization (6 tasks: circle packing, Erd\H{o}s overlap, signal processing, autocorrelation inequalities, min-max distance), systems optimization (5 tasks: EPLB, PRISM, LLM-SQL, transaction scheduling, Cloudcast), and stress-test problems (Kernel Engineering, Polyominoes). Grading approaches include subprocess execution with JSON result parsing, constraint validation, benchmark-relative scoring, and delegation to external evaluation frameworks.

\subsection{Evaluator Corrections}
\label{app:eval_corrections}

The systems optimization tasks in our evaluation follow the ADRS folder in the SkyDiscover repository~\citep{skydiscover2026}, which provides evaluator implementations originally from the Sky-Discover repository. During integration and testing, we discovered and corrected several bugs in these evaluators that could lead to incorrect scoring. We document all corrections here for reproducibility; all fixes are available in our public repository. Table~\ref{tab:eval_fixes} summarizes the corrections, and we describe each in detail below.

\begin{table}[t]
\caption{Evaluator bug fixes applied to the ADRS benchmark evaluators. All bugs were present in the original implementations; our fixes ensure correct scoring.}
\label{tab:eval_fixes}
\centering
\scriptsize
\setlength{\tabcolsep}{8pt}
\begin{tabular}{l p{5cm} p{6cm}}
\toprule
\textbf{Task} & \textbf{Bug} & \textbf{Fix} \\
\midrule
PRISM & Failed placements silently skipped & Append worst-case penalty (${10}^{6}$) for failures \\
Txn Sched. & Invalid schedules scored $>0$ & Return score 0 for invalid schedules \\
EPLB & Experts with 0 replicas skipped; 3-run averaging masked issues & Penalize by concentrating load on slot~0; remove redundant averaging \\
LLM-SQL & Mixed-type DataFrame crashes & Convert all values to string dtype \\
\bottomrule
\end{tabular}
\end{table}

\paragraph{PRISM: failed placements silently skipped.}
The original PRISM evaluator catches \texttt{TimeoutError} and general exceptions during GPU placement evaluation but simply calls \texttt{continue}, skipping the failed test case entirely. This means a solution that crashes on difficult inputs can achieve an artificially high average score by only being graded on easy cases. Our fix appends a worst-case penalty value ($10^6$, corresponding to maximum load imbalance) for each failed placement, ensuring that failures are reflected in the final score.

\paragraph{Transaction Scheduling: invalid schedules scored above zero.}
The transaction scheduling evaluator computes a score via $\text{score} = 10^6 / (1 + \text{makespan})$ regardless of whether the schedule is valid (i.e., respects read-write and write-write conflict ordering). Invalid schedules thus received positive scores proportional to their makespan, rewarding incorrect solutions. Our fix gates the formula on a validity check: invalid schedules receive a score of 0.

\paragraph{EPLB: dropped experts and redundant averaging.}
The EPLB (Expert-Parallel Load Balancing) evaluator had two issues. First, when an expert had zero replicas assigned, the evaluator skipped it instead of penalizing the imbalance. This allowed solutions that simply drop difficult-to-balance experts to appear well-balanced. Our fix concentrates all of the dropped expert's load onto a single physical slot (the worst-case imbalance). Second, the grader averaged results over 3 redundant runs of the same deterministic evaluator, which added noise without value. We removed this averaging and use a single evaluation.

\paragraph{LLM-SQL: type handling.}
 The seed program column analysis assumed homogeneous column types in the input DataFrame, but real datasets contain mixed types (integers, strings, nulls) that cause prefix-matching to crash. Our fixes convert all DataFrame values to string dtype before analysis.

\section{Experimental Details}
\label{app:experiments}

\subsection{Setup Details}
\label{app:setup}

\paragraph{Hardware.} All experiments were conducted on Linux machines. Mathematical and systems optimization tasks were run on CPU-only instances, as these tasks do not require GPU acceleration. The Kernel Engineering and Polyominoes stress-test tasks were similarly run on CPU instances, as evaluation involves simulation rather than GPU computation.

\paragraph{Models.} Our primary backbone model is Claude Opus 4.6 (\texttt{claude-opus-4-6}), used for both \method{} agents and all baselines (OpenEvolve, ShinkaEvolve, EvoX). For the open-source generalization experiments, we use MiniMax M2.5 with the OpenCode~\citep{opencode2025} runtime. No internet access is provided to agents unless the task configuration explicitly enables the \texttt{research} flag.

\paragraph{Baselines.} We compare against three fixed evolutionary search baselines:
\begin{itemize}[nosep,leftmargin=*]
    \item \textbf{OpenEvolve}~\citep{sharma2025openevolve}: Open-source implementation of AlphaEvolve with static elite populations and diversity maintenance.
    \item \textbf{ShinkaEvolve}~\citep{lange2025shinkaevolve}: Adaptive sampling with bandit-based selection.
    \item \textbf{EvoX}~\citep{liu2026evox}: Meta-evolved search strategy with co-evolutionary outer loop.
\end{itemize}
All baselines receive identical seed programs, evaluators, and wall-clock budgets. For the mathematical and systems suites, we follow the protocol defined in the SkyDiscover GitHub repository~\citep{skydiscover2026}.

\paragraph{Evaluation protocol.} For the mathematical and systems optimization suites (Table~\ref{tab:main_results}), all methods are given a 3-hour wall-clock budget, averaged over 4 independent runs. For the stress-test problems (Table~\ref{tab:multi-agent}), experiments terminate when there is no improvement over 100 evaluations or 2 hours, whichever comes first. Multi-agent experiments use 4 agents with matched wall-clock time. The single-agent Bo4 baseline reports the best score across 4 independent single-agent runs, approximating $4\times$ compute without multi-agent coordination.

\paragraph{Configuration.} Unless otherwise noted, all \method{} experiments use the default heartbeat configuration (Table~\ref{tab:heartbeat_config}). Agents are configured with \texttt{max\_turns=200}. Agents are initialized identically in multi-agent experiments (no role specialization).

\subsection{Cost and Efficiency}
\label{app:cost}

\paragraph{API costs.} For a typical 3-hour single-agent run on a mathematical optimization task using Claude Opus 4.6, total API cost ranges from approximately \$30--60 USD depending on task complexity. Multi-agent runs with 4 agents incur approximately $3$--$4\times$ the single-agent cost. Context caching in the Claude API reduces costs for repeated context windows within a session.

\paragraph{Evaluation efficiency.} \method{} agents typically perform fewer evaluation calls than structured baselines within the same wall-clock budget, because each agent step involves reasoning and implementation before submission. However, the improvement rate (fraction of evaluations yielding a score improvement) is substantially higher (Table~\ref{tab:main_results}), indicating more efficient use of each evaluation call.

\paragraph{Infrastructure overhead.} The \method{} infrastructure (manager, monitoring loop, file watching) adds negligible overhead---the monitoring loop polls every 5 seconds and heartbeat prompt rendering is instantaneous. The dominant costs are LLM API calls and grader execution.

\end{document}